%% file: main.tex
\documentclass{article}

\usepackage[margin=1in]{geometry}
\usepackage{mathpazo}

\usepackage[T1]{fontenc}
\usepackage{amsmath,amssymb,amsthm}
\usepackage{hyperref}
\usepackage{graphicx}
\usepackage{color}
\usepackage{booktabs}
\usepackage{multirow}
\usepackage{subcaption}
\usepackage{makecell}
\usepackage{adjustbox}

\usepackage{listings}
\definecolor{mydarkblue}{rgb}{0,0.08,0.85}
\hypersetup{colorlinks=true,
linkcolor=mydarkblue,
citecolor=mydarkblue,
filecolor=mydarkblue,
urlcolor=mydarkblue,
bookmarksopen=true}

\usepackage[backend=biber,style=alphabetic,natbib=true,maxnames=99,maxcitenames=2,minalphanames=3]{biblatex}
\title{Do Adversarially Robust ImageNet \\ Models Transfer Better?} 
\author{
    Hadi Salman\footnote{Equal contribution.} \\
    \texttt{hadi.salman@microsoft.com} \\
    Microsoft Research
    \and
    Andrew Ilyas\footnotemark[1] \\
    \texttt{ailyas@mit.edu} \\
    MIT
    \and 
    Logan Engstrom \\
    \texttt{engstrom@mit.edu} \\
    MIT
    \and 
    Ashish Kapoor \\
    \texttt{akapoor@microsoft.com} \\
    Microsoft Research
    \and
    Aleksander M\k{a}dry \\
    \texttt{madry@mit.edu} \\
    MIT
}
\date{}

\begin{document}
    \maketitle

    \begin{abstract}
        \input{sections/abstract}
    \end{abstract}

    \section{Introduction}
    \label{sec:intro}
    \input{sections/intro}

    \section{Motivation: Fixed-Feature Transfer Learning}
    \label{sec:motivation}
    \input{sections/fixed_feature}

    \section{Adversarial Robustness and Full-Network Fine Tuning}
    \label{sec:fullnetwork}
    \input{sections/full_network}

    \section{Analysis and Discussion}
    \label{sec:analysis}
    \input{sections/analysis}

    \section{Related Work}
    \input{sections/related}

    \section{Conclusion}
    \input{sections/conclusion}

    \section*{Acknowledgements}
    \input{sections/acknowledgements}

    \clearpage
    \printbibliography
    \clearpage
    \appendix
    \section{Experimental Setup}
    \label{app:experimental-setup}
    \input{sections/app_experimental_setup}

    \clearpage
    \input{sections/app_rel}
    \input{sections/app-background}
    
    \clearpage
    \input{sections/app-omitted-figures}

    \clearpage
    \section{Detailed Numerical Results}
    \label{app:numerical-results}
    \input{sections/app_numerical_results}

    \end{document}

%% file: sections/abstract.tex
Transfer learning is a widely-used paradigm in which models
pre-trained on standard datasets can efficiently adapt to downstream tasks.
Typically,
better pre-trained models yield better transfer results, suggesting that
initial accuracy is a key aspect of transfer learning performance. 
In this work, we identify another such aspect: we find that adversarially robust
models, while less accurate, often perform better than their standard-trained
counterparts when used for transfer learning. 
Specifically, we focus on adversarially robust ImageNet
classifiers, and show that they yield improved accuracy on a standard suite of
downstream classification tasks.  
Further analysis 
uncovers more differences between robust and standard models in the
context of transfer learning.
Our results are consistent with (and in fact, add to) recent hypotheses
stating that robustness leads to improved feature representations. 
Our code and models are available at 
\url{https://github.com/Microsoft/robust-models-transfer}.

%% file: sections/intro.tex
Deep neural networks currently define state-of-the-art performance across many
computer vision tasks.
When large quantities of labeled data and computing resources are available, models perform well when trained from scratch.
However, in many practical settings there is insufficient data or 
compute for this approach to be viable. 
In these cases, {\em transfer learning}~\citep{donahue2014decaf, razavian2014cnn} has emerged as a simple and
efficient way to obtain performant models. 
Broadly, transfer learning refers to any machine learning algorithm
that leverages information from one (``source'') task to better solve another
(``target'') task. 
A prototypical transfer learning pipeline in computer vision (and the focus of
our work) starts with a model trained on the ImageNet-1K
dataset~\citep{deng2009imagenet, russakovsky2015imagenet}, and then refines this
model for the target task.

Though the exact underpinnings of transfer learning are not fully understood, 
recent work has identified factors that make pre-trained ImageNet models amenable to 
transfer learning. 
For example,~\citep{huh2016makes, kolesnikov2019big} investigate the effect of
the source dataset; \citet{kornblith2019better} find that pre-trained models
with higher ImageNet accuracy also tend to transfer
better;~\citet{azizpour2015factors} observe that increasing depth improves
transfer more than increasing width.

\paragraph{Our contributions.}
In this work, we identify another factor that affects transfer learning performance:
adversarial robustness~\citep{biggio2013evasion,szegedy2014intriguing}.
We find that despite being less accurate on ImageNet, adversarially robust
neural networks match or improve on the transfer performance of their standard
counterparts. 
We first establish this trend in the ``fixed-feature'' setting,
in which one trains a linear classifier on top of features extracted from a 
pre-trained network.
Then, we show that this trend carries forward to the more complex
``full-network'' transfer setting, in which the pre-trained model is entirely
fine-tuned on the relevant downstream task.   
We carry out our study on a suite of image classification tasks (summarized in
Table~\ref{tab:headline}), object detection, and instance segmentation.

Our results are consistent with (and in fact, add to)
recent hypotheses suggesting that adversarial robustness leads to improved
feature representations~\citep{engstrom2019learning, allenzhu2020feature}.
Still, future work is needed to confirm or refute such hypotheses, and more
broadly, to understand what properties of pre-trained models are important for
transfer learning. 

\renewcommand{\rothead}[2][60]{\makebox[9mm][c]{\rotatebox{#1}{\makecell[c]{#2}}}}%
\begin{table}
    \centering
    \caption{Transfer performance of robust and standard ImageNet
    models on 12 downstream classification tasks. For each transfer
    learning paradigm, we report accuracy averaged over ten random trials of
    standard and robust models. We used a grid search (using a disjoint set of
    random seets) to find the best hyperparameters, architecture, and (for
    robust models) robustness level $\varepsilon$.
    In each column, we bold the entry with the higher average accuracy; if the
    accuracy difference is significant (as judged by a 95\% CI two-tailed
    Welch's t-test \citep{welch1947generalization}) we bold only the higher entry, otherwise (if the test is
    inconclusive) we bold both.} 
    \label{tab:headline}
\resizebox{\linewidth}{!}{
    \begin{tabular}{@{}llcccccccccccc@{}}
    & {} & \multicolumn{12}{c}{Dataset} \\
    \cmidrule(lr){3-14}
    \textbf{Mode} & \textbf{Model} &              \rothead{Aircraft} & \rothead{Birdsnap} &
    \rothead{CIFAR-10} & \rothead{CIFAR-100} & 
    \rothead{Caltech-101} & \rothead{Caltech-256} &  
    \rothead{Cars} & \rothead{DTD} & \rothead{Flowers} &  
    \rothead{Food} &  \rothead{Pets} & \rothead{SUN397} \\
    \midrule
    \textbf{Fixed-} & \textbf{Robust} & {\bf 44.24} &    {\bf 50.75} &    {\bf 95.50} &  {\bf 81.16} &   {\bf 92.54} &     {\bf 85.16} & {\bf 51.35} & {\bf 70.38} &   {\bf 92.05} & {\bf 69.32} & {\bf 92.08} & \textbf{58.80} \\ 
    \textbf{feature} & \textbf{Standard} &                 38.52       &         48.40  &    81.29       &     60.08    &       90.01 &        82.87     & 44.54 & \textbf{70.32} &   91.83 & 65.73 & 91.92 &  56.02 \\
    \midrule
    \textbf{Full-} & {\bf  Robust} & \textbf{86.26} & {\bf     76.41 } & {\bf     98.70 } & {\bf      89.22 } & {\bf        95.67 } & {\bf        87.92 } & \textbf{91.37} & {\bf 77.05 } & {\bf    96.94 } & {\bf  89.10 } & {\bf  94.36 } & {\bf   64.97} \\
    \textbf{network} & \textbf{Standard} &  {\bf 86.19} &   75.90 &    97.72 &     86.20 & 94.85 &       86.54 & {\bf 91.37} & 76.11 &   \textbf{97.13} & 88.61 & \textbf{94.43} &  63.90 \\
    \bottomrule
    \end{tabular}
    }
\end{table}

%% file: sections/fixed_feature.tex
In one of the most basic variants of transfer learning, one uses the source
model as a feature extractor for the target dataset, then trains a simple
(often linear) model on the resulting features.
In our setting, this corresponds to first passing each image in the target
dataset through a pre-trained ImageNet classifier, and then using the outputs
from the penultimate layer as the image's feature representation.
Prior work has demonstrated that applying this ``fixed-feature'' transfer
learning approach yields accurate classifiers for a variety of vision tasks and 
often out-performs task-specific handcrafted features~\citep{razavian2014cnn}. 
However, we still do not completely understand the factors driving transfer
learning performance.  

\paragraph{How can we improve transfer learning?}
Conventional wisdom and evidence from prior work
\citep{chatfield2014return, simonyan2015very, kornblith2019better,
huang2017speed} suggest that accuracy on the source dataset is a strong
indicator of performance on downstream tasks.
In particular,~\citet{kornblith2019better} find that pre-trained ImageNet models with
higher accuracy yield better fixed-feature transfer learning results.

Still, it is unclear if improving ImageNet accuracy is the only way to improve
performance. 
After all, the behaviour of fixed-feature transfer is governed by
models' learned representations, which are not fully described by source-dataset
accuracy. 
These representations are, in turn, controlled by the {\em priors} that we put on
them during training. 
For example, the use of architectural components~\citep{ulyanov2017deep},
alternative loss functions~\citep{muralidhar2018incorporating}, and data
augmentation~\citep{van2001art} have all been found to put distinct priors on
the features extracted by classifiers. 

\paragraph{The adversarial robustness prior.}
In this work, we turn our attention to another prior: {\em adversarial
robustness}.
Adversarial robustness refers to a model's invariance to small (often
imperceptible) perturbations of its inputs. 
Robustness is typically induced at training time by replacing the standard
empirical risk minimization objective with a robust optimization
objective~\citep{madry2018towards}:
\begin{equation}
    \label{eq:robustobjective}
    \min_{\theta} \mathbb{E}_{(x,y)\sim D}\left[\mathcal{L}(x,y;\theta)\right]
    \implies
    \min_{\theta} \mathbb{E}_{(x,y)\sim D}
    \left[\max_{\|\delta\|_2 \leq \varepsilon} \mathcal{L}(x+\delta,y;\theta) \right],
\end{equation}
where $\varepsilon$ is a hyperparameter governing how invariant the resulting
``adversarially robust model'' (more briefly, ``robust model'') should be. 
In short, this objective asks the model to minimize risk on the 
training datapoints while also being locally stable in the (radius-$\varepsilon$)
neighbourhood around each of these points. (A more detailed primer on
adversarial robustness is given in Appendix~\ref{app:background}.)

Adversarial robustness was originally studied in the context of machine learning
security~\citep{biggio2013evasion, biggio2018wild,
carlini2017adversarial} as a method for improving models'
resilience to adversarial examples \citep{goodfellow2015explaining,madry2018towards}.
However, a recent line of work has studied adversarially robust models in their 
own right, casting~\eqref{eq:robustobjective} as a prior on
learned feature representations~\citep{engstrom2019learning, ilyas2019adversarial,  
jacobsen2019excessive, zhang2019interpreting}. 

\paragraph{Should adversarial robustness help fixed-feature transfer?} 
It is, a priori, unclear what to expect from an ``adversarial robustness
prior'' in terms of transfer learning. 
On one hand, robustness to adversarial examples may seem somewhat tangential
to transfer performance.
In fact, adversarially robust models are known to be significantly less
accurate than their standard counterparts \citep{tsipras2019robustness,
su2018robustness, raghunathan2019adversarial, nakkiran2019adversarial},
suggesting that using adversarially robust feature
representations should hurt transfer performance. 

\begin{figure}[htbp]
    \centering
    \begin{subfigure}[]{0.48\linewidth}
        \centering
        \includegraphics[width=\linewidth]{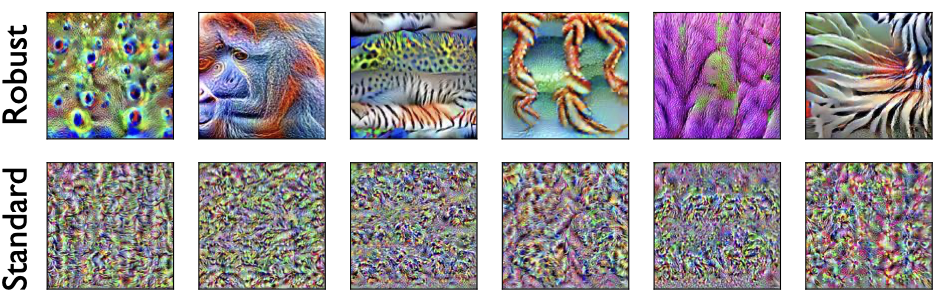}
        \caption{Perceptually aligned gradients}
        \label{fig:feature_viz}
    \end{subfigure}
    \hspace{3em}
    \begin{subfigure}[]{0.34\linewidth}
        \centering
        \includegraphics[width=\linewidth]{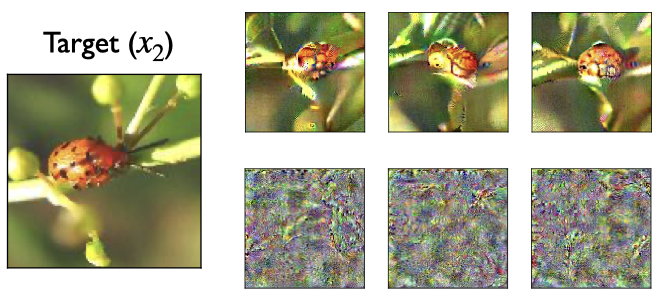}
        \caption{Representation invertibility}
        \label{fig:invertibility}
    \end{subfigure}
    \caption{Adversarially robust (top) and standard (bottom) representations:
    robust representations allow (a) feature visualization without
    regularization; (b) approximate image inversion by minimizing distance in
    representation space. Figures reproduced from \citet{engstrom2019learning}.} 
\end{figure}

On the other hand, recent work has found that the feature representations of
robust models carry several advantages over those of standard models.  
For example, adversarially robust representations typically have better-behaved
gradients~\citep{tsipras2019robustness, santurkar2019image,
zhang2019interpreting, kaur2019perceptually} and thus facilitate
regularization-free feature visualization~\citep{engstrom2019learning} (cf.
Figure~\ref{fig:feature_viz}).   
Robust representations are also approximately
invertible~\citep{engstrom2019learning}, meaning that unlike for standard
models~\citep{mahendran2015understanding, dosovitskiy2016inverting}, an image
can be approximately reconstructed directly from its robust representation (cf. 
Figure~\ref{fig:invertibility}).
More broadly,~\citet{engstrom2019learning} hypothesize that by forcing networks
to be invariant to signals that humans are also invariant to, the robust
training objective leads to feature representations that are more similar to
what humans use.  
This suggests, in turn, that adversarial robustness might be a desirable
prior from the point of view of transfer learning.

\paragraph{Experiments.} To resolve these two conflicting hypotheses, we use a
test bed of 12 standard transfer learning datasets (all the datasets considered
in~\citep{kornblith2019better} as well as Caltech-256~\citep{griffin2007caltech}) to evaluate fixed-feature
transfer on standard and adversarially robust ImageNet models. We considere
four ResNet-based architectures (ResNet-\{18,50\}, WideResNet-50-x\{2,4\}), and
train models with varying robustness levels $\varepsilon$ for each architecture
(for the full experimental setup, see Appendix~\ref{app:experimental-setup}).  

In Figure~\ref{fig:fixed_feature_bar}, we compare the downstream transfer
accuracy of a standard model to that of the best robust model
with the same architecture (grid searching over $\varepsilon$\footnote{To ensure
a fair comparison (i.e., that the gains observed are not an artifact of training
many random robust models), we first use a set of random seeds to select the best
$\varepsilon$ level, and then calculate the perfomance for just that
$\varepsilon$ using a separate set of random seeds.}).
The results indicate that robust networks consistently extract better features
for transfer learning than standard networks---this effect is most pronounced on
Aircraft, CIFAR-10, CIFAR-100, Food, SUN397, and Caltech-101. Due to computational
constraints, we could not train WideResNet-50-4x models at the same number of
robustness levels $\varepsilon$, so a coarser grid was used. It is thus likely
that a finer grid search over $\varepsilon$ would further improve results 
(we discuss the role of $\varepsilon$ in more detail in
Section~\ref{sec:effective-epsilon-hypothesis}).    

\begin{figure}[!htbp]
    \centering
    \includegraphics[width=.9\linewidth]{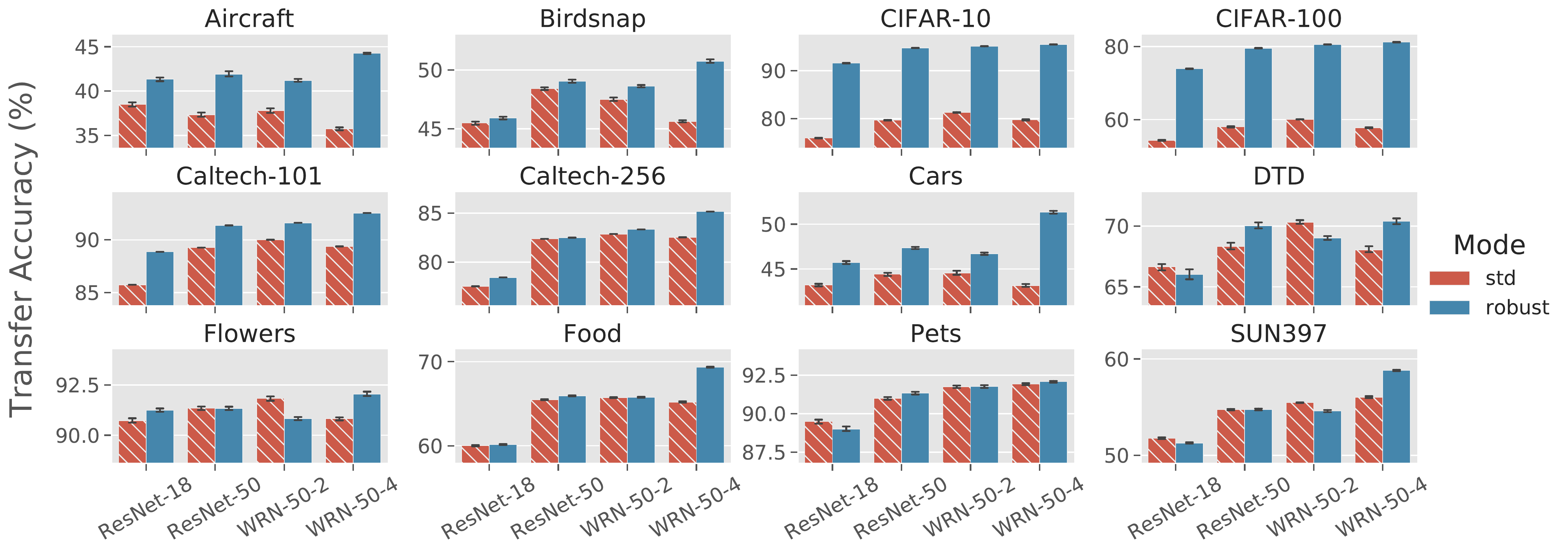}
    \caption{\textbf{Fixed-feature} transfer learning results using standard and
    robust models for the 12 downstream image classification tasks
    considered. Following~\citep{kornblith2019better}, we record re-weighted
    accuracy for the unbalanced datasets, and raw accuracy for the others (cf.
    Appendix~\ref{app:experimental-setup}). Error bars denote one standard
    deviation computed over ten random trials.} 
    \label{fig:fixed_feature_bar}
\end{figure}

%% file: sections/full_network.tex
A more expensive but often better-performing transfer learning method uses
the pre-trained model as a weight initialization rather than as a feature
extractor.
In this ``full-network'' transfer learning setting, we update all of the weights
of the pre-trained model (via gradient descent) to minimize loss on the target
task. 
\citet{kornblith2019better} find that for standard models, performance on
full-network transfer learning is highly correlated with performance on
fixed-feature transfer learning.
Therefore, we might hope that the findings of the last section (i.e., that
adversarially robust models transfer better) also carry over to this setting. 
To resolve this conjecture, we consider three applications of full-network
transfer learning: image classification (i.e., the tasks considered
in Section~\ref{sec:motivation}), object detection, and instance segmentation.

\begin{figure}[!h]
    \centering
    \includegraphics[width=.9\linewidth]{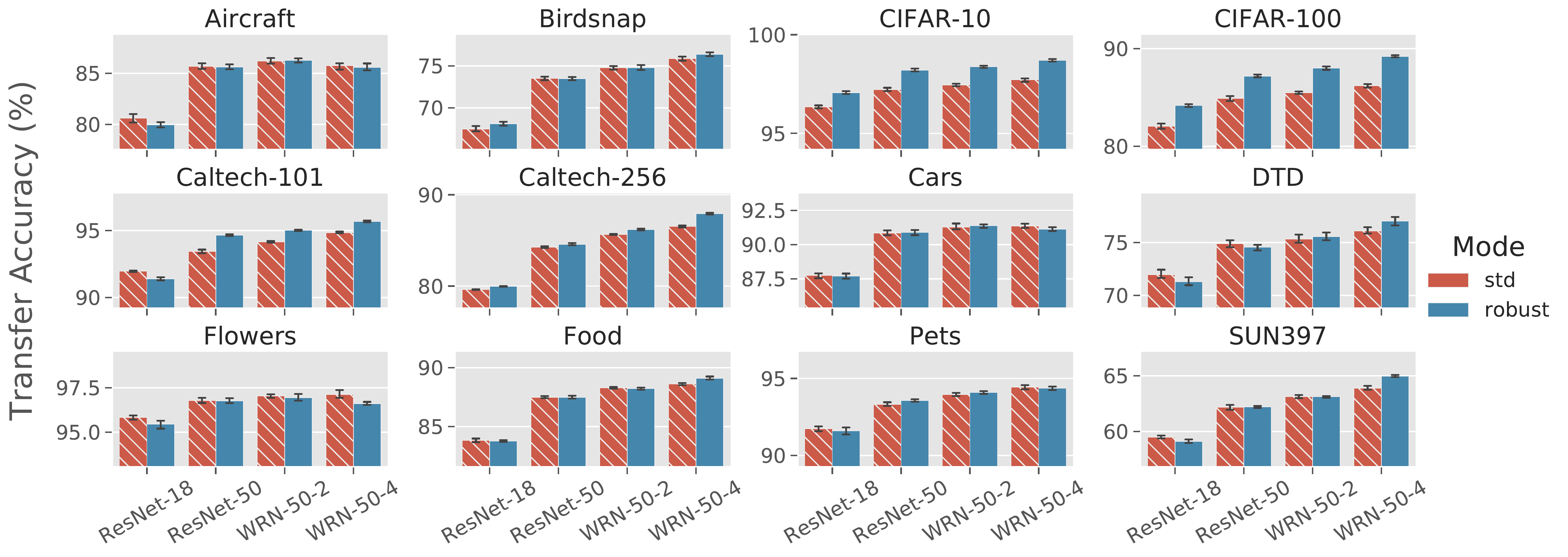}
    \caption{\textbf{Full-network} transfer learning results using standard and
    robust models for the 12 downstream image classification tasks
    considered. Following~\citep{kornblith2019better}, we record re-weighted
    accuracy for the unbalanced datasets, and raw accuracy for the others (cf.
    Appendix~\ref{app:experimental-setup}). Error bars denote one standard
    deviation computed with ten random trials.}
    \label{fig:main-small-datasets-transfer-results-finetuning}
\end{figure}

\paragraph{Downstream image classification}
We first recreate the setup of Section~\ref{sec:motivation}: we
perform full-network transfer learning to adapt the robust and non-robust
pre-trained ImageNet models to the same set of 12 downstream classification
tasks. 
The hyperparameters for training were found via grid search (cf.
Appendix~\ref{app:experimental-setup}).  
Our findings are shown in
Figure~\ref{fig:main-small-datasets-transfer-results-finetuning}---just as in
fixed-feature transfer learning, robust models match or improve on standard
models in terms of transfer learning performance. 

\input{sections/objdet_instseg.tex}

%% file: sections/objdet_instseg.tex
\paragraph{Object detection and instance segmentation}
It is standard practice in
data-scarce object detection or instance segmentation tasks to initialize
earlier model layers with weights from ImageNet-trained classification networks.
We study the benefits of using robustly trained networks to initialize object
detection and instance segmentation models, and find that adversarially robust
networks consistently outperform standard networks.

We evaluate with benchmarks in both object detection (PASCAL Visual
Object Classes (VOC)~\citep{everingham2010pascal} and Microsoft
COCO~\citep{lin2014microsoft}) and instance segmentation (Microsoft COCO). We
train systems using default models and hyperparameter configurations from the
Detectron2~\citep{wu2019detectron2} framework (i.e., we do not perform any
additional hyperparameter search). Appendix~\ref{app:det} describes further
experimental details and more results.

\begin{figure}
\centering
\includegraphics[width=.95\textwidth]{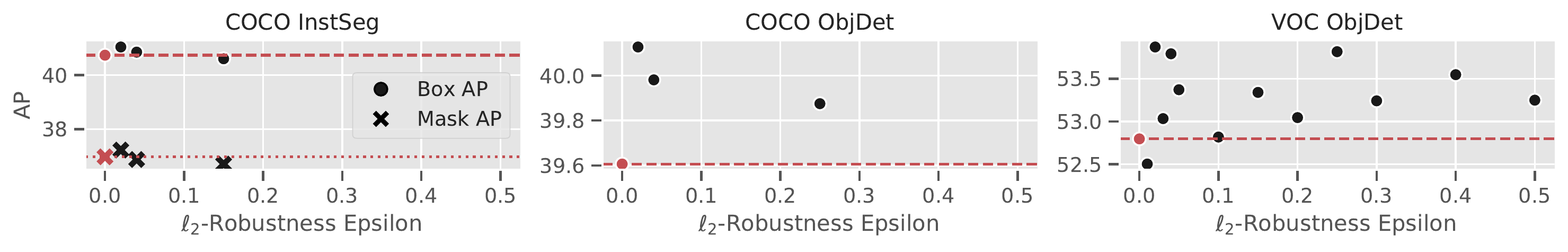}
\vspace{1em}
\small
\begin{tabular}{lrrrr}
\toprule
\multirow{2}{*}{\textbf{Task}} &  \multicolumn{2}{c}{\textbf{Box AP}} & \multicolumn{2}{c}{\textbf{Mask AP}} \\ \cmidrule(lr){2-3} \cmidrule(lr){4-5}
                        & Standard & Robust & Standard & Robust \\
\midrule
       VOC Object Detection &            $52.80$ &          $53.87$ &               --- &             --- \\
      COCO Object Detection &            $39.80\pm0.08$ &       $40.07\pm0.10$ &               --- &             --- \\
 COCO Instance Segmentation &            $40.67\pm0.06$ &          $40.91\pm0.15$ &             $36.92\pm0.08$&           $37.08\pm0.10$ \\
\bottomrule
\end{tabular}
 \caption{AP of instance segmentation and object detection models with backbones
   initialized with $\varepsilon$-robust models before training. Robust
   backbones generally lead to better AP, and the best robust backbone always
   outperforms the standardly trained backbone for every task. COCO results
   averaged over four runs due to computational constraints; $\pm$
   represents standard deviation.} 
\label{fig:eps_vs_ap_3x}
\end{figure}

We first study object detection. We train Faster R-CNN
FPN~\citep{lin2017feature} models with 
varying ResNet-50 backbone initializations.
For VOC, we initialize with one
standard network, and twelve adversarially robust networks with different
values of $\varepsilon$. 
For COCO, we only train with three adversarially robust
models (due to computational constraints). 
For instance segmentation, we train Mask R-CNN FPN
models~\citep{he2017mask} while varying ResNet-50 backbone initialization. We
train three models using adversarially robust initializations, and one model
from a standardly trained ResNet-50. Figure~\ref{fig:eps_vs_ap_3x} summarizes our
findings: the best robust backbone initializations outperform standard models.

%% file: sections/analysis.tex
Our results from the previous section indicate that robust models match or
improve on the transfer learning performance of standard ones.
In this section, we take a closer look at the similarities and differences in
transfer learning between robust networks and standard networks.

\input{sections/accuracy_vs_transfer}

\input{sections/width_effect}

\input{sections/effective_epsilon}

\input{sections/stylized_comparison}

%% file: sections/accuracy_vs_transfer.tex
\subsection{ImageNet accuracy and transfer performance}
In Section~\ref{sec:motivation}, we discussed a potential tension between the
desirable properties of robust network representations 
(which we conjectured would improve transfer performance) and the decreased
accuracy of the corresponding models (which, as prior work has established,
should hurt transfer performance).
We hypothesize that robustness and accuracy have counteracting yet
separate effects: that is, higher accuracy improves transfer
learning for a fixed level of robustness, and higher robustness improves transfer learning for a fixed level of accuracy.

To test this hypothesis, we first study the relationship between ImageNet
accuracy and transfer accuracy for each of the robust models that we trained.
Under our hypothesis, we should expect to see a deviation from the direct linear
accuracy-transfer relation observed by~\citep{kornblith2019better}, due to the
confounding factor of varying robustness.
The results (cf.
Figure~\ref{fig:main-small-datasets-transfer-results-logistic-regression};
similar results for full-network transfer in Appendix~\ref{app:omitted-results})
support this.
Indeed, we find that the previously observed linear relationship between accuracy and transfer
performance is often violated once robustness aspect comes into play.

\begin{figure}[t]
        \begin{subfigure}[]{\linewidth}
            \centering
            \includegraphics[width=0.33\linewidth]{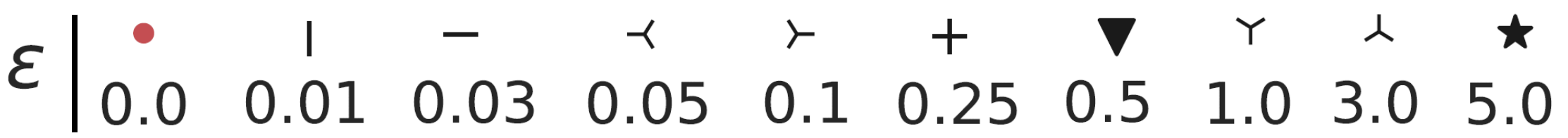}
        \end{subfigure}
        \begin{subfigure}[]{0.5\linewidth}
            \adjincludegraphics[height=4.6cm,trim={{.005\width} 0 {.09\width} 0},clip]{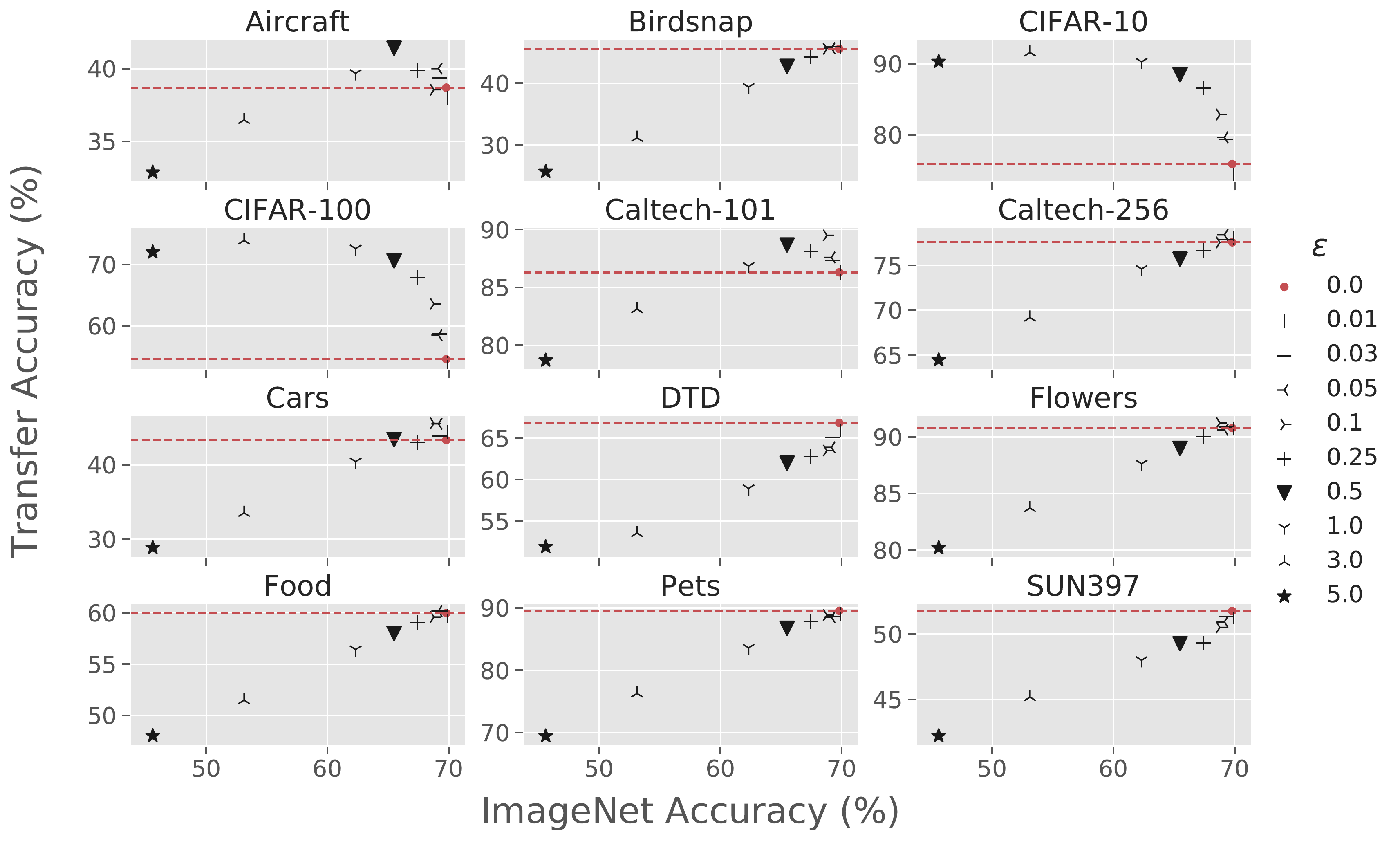}
            \caption{ResNet-18}
        \end{subfigure}
        \begin{subfigure}[]{0.5\linewidth}
            \adjincludegraphics[height=4.6cm,trim={{.035\width} 0 {.09\width} 0},clip]{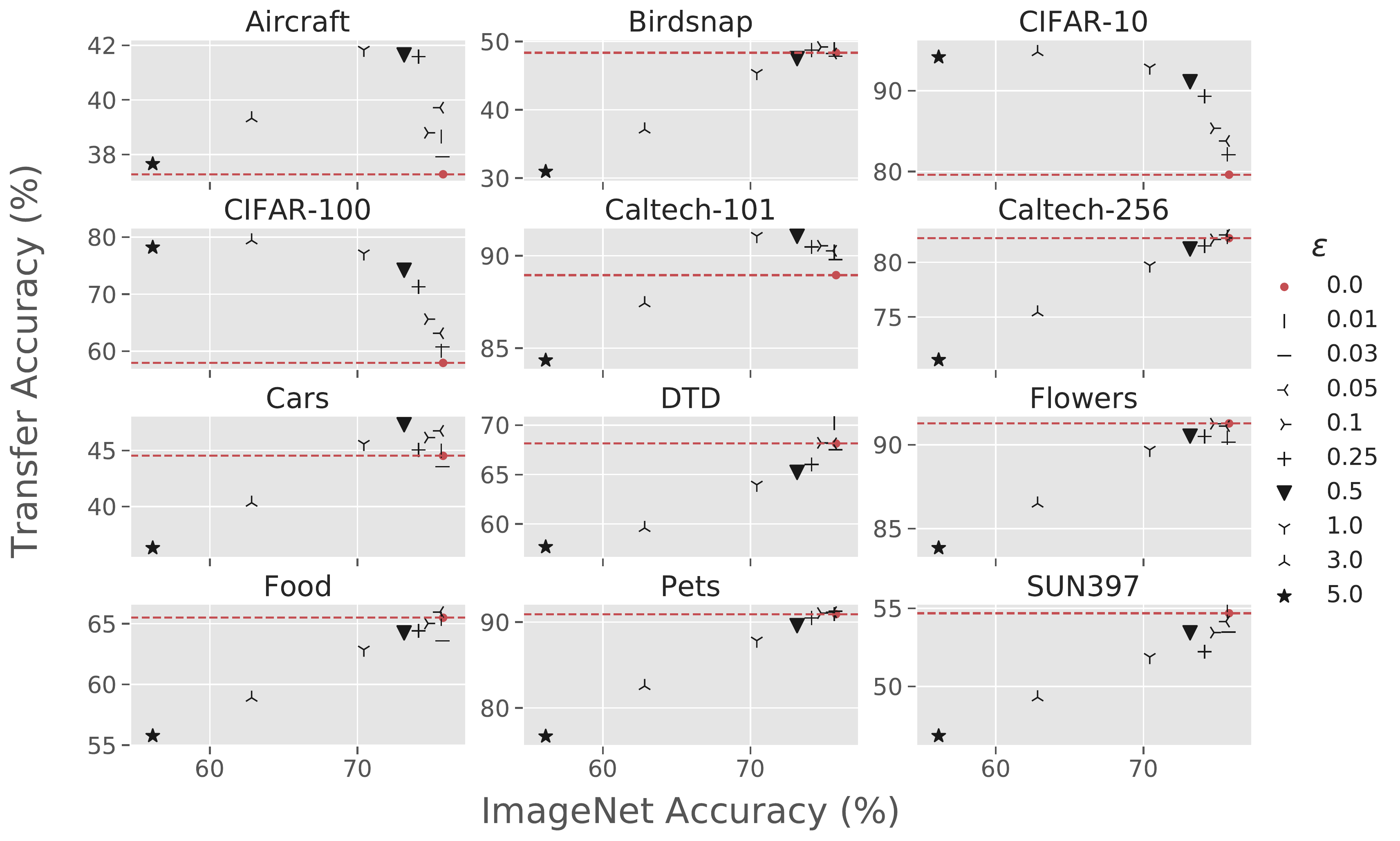}
            \caption{ResNet-50}
        \end{subfigure}
        \begin{subfigure}[]{0.5\linewidth}
            \adjincludegraphics[height=4.6cm,trim={{.005\width} 0 {.09\width} 0},clip]{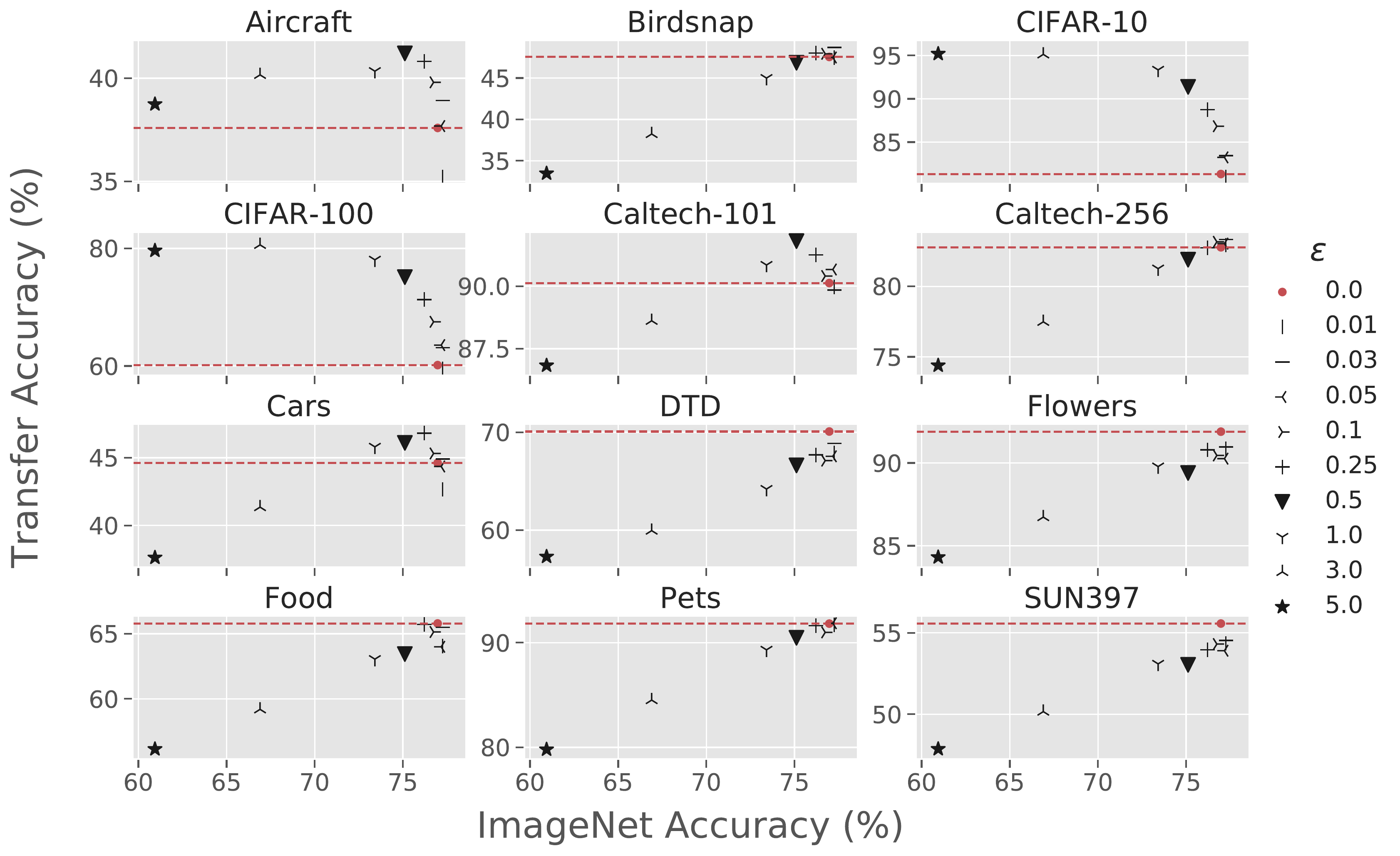}
            \caption{WRN-50-2}
        \end{subfigure}
        \begin{subfigure}[]{0.5\linewidth}
            \adjincludegraphics[height=4.6cm,trim={{.035\width} 0 {.09\width} 0},clip]{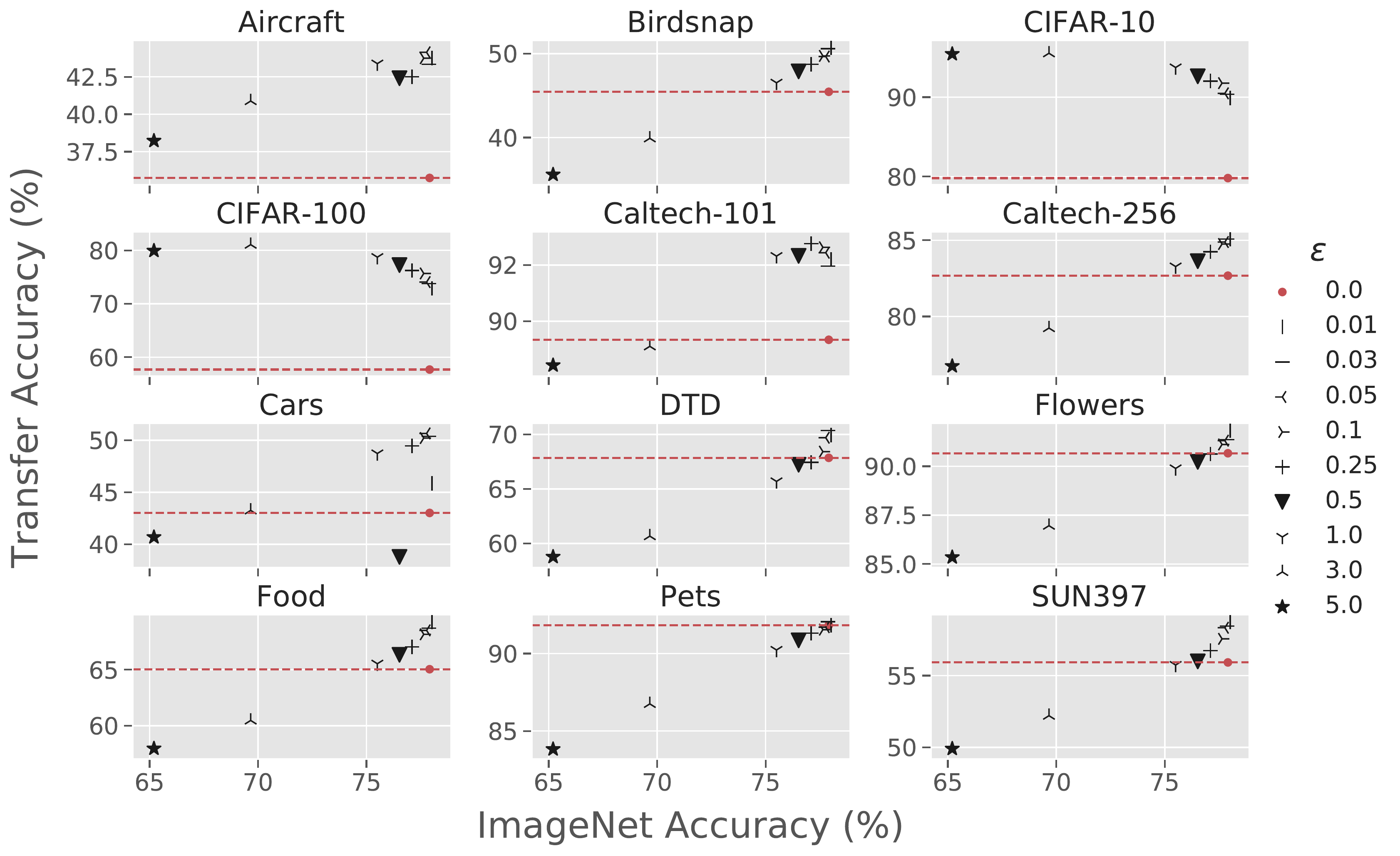}
            \caption{WRN-50-4}
        \end{subfigure}
     \caption{\textbf{Fixed-feature} transfer accuracies of standard and robust
       ImageNet models to various image classification datasets. The linear
       relationship between ImageNet and transfer accuracies does not hold. Full
       numerical results (i.e., in tabular form) are available in
       Appendix~\ref{app:numerical-results}.}
     \label{fig:main-small-datasets-transfer-results-logistic-regression}
\end{figure}

\begin{table}[t]
\centering
\caption{Source (ImageNet) and target (CIFAR-10) accuracies, fixing robustness
($\varepsilon$) but varying architecture. When robustness is controlled for,
ImageNet accuracy is highly predictive of transfer performance. Similar trends
for other datasets are shown in Appendix~\ref{app:omitted-results}.}
\label{tab:linear_relation}
\begin{tabular}{@{}llcccccc@{}cc@{}}
    \toprule
    & & \multicolumn{6}{c}{\textbf{Architecture} (see details in Appendix \ref{app:pretrained-models})} & & \\
    \cmidrule(lr){3-9}
    \textbf{Robustness} & \textbf{Dataset} &  A &  B &  C &  D &  E &  F & &
    $R^2$ \\
    \midrule
    \multirow{2}{*}{Std ($\varepsilon=0$)  } & ImageNet &    77.37 & 77.32 &
    73.66 & 65.26 & 64.25 & 60.97 && --- \\ 
    & CIFAR-10 &         97.84 &  97.47  &        96.08 &     95.86 &      95.82
    &    95.55 & & 0.79 \\
    \midrule
    \multirow{2}{*}{Adv ($\varepsilon=3$)  } & ImageNet &        66.12 & 65.92 &
    56.78 & 50.05 & 42.87 & 41.03 && --- \\
    & CIFAR-10 &  98.67 & 98.22 &   97.27 &  96.91 & 96.23 &  95.99 && 0.97 \\
    \bottomrule
\end{tabular}    
\end{table}

In even more direct support of our hypothesis (i.e., that robustness and
ImageNet accuracy have opposing yet separate effects on transfer), we
find that when the robustness level is held fixed, the accuracy-transfer
correlation observed by prior works for standard models actually holds for
robust models too. 
Specifically, we train highly robust ($\varepsilon = 3$)---and
thus less accurate---models with six different architectures, and compared
ImageNet accuracy against transfer learning performance. 
Table~\ref{tab:linear_relation} shows that for these models
improving ImageNet accuracy improves transfer performance at around the same
rate as (and with higher $R^2$ correlation than) standard models. 

These observations suggest that
transfer learning performance can be further improved by applying known
techniques that increase the accuracy of robust models
(e.g.~\citep{balaji2019instance, carmon2019unlabeled}).
More broadly, our findings also indicate that accuracy is not a sufficient
measure of feature quality or versatility. 
Understanding why robust networks transfer particularly well remains an open
problem, likely relating to prior work that analyses the 
features these networks use~\citep{engstrom2019learning,
shafahi2019adversarially, allenzhu2020feature}.

%% file: sections/width_effect.tex
\subsection{Robust models improve with width}
\begin{figure}[!htbp]
    \centering
    \begin{subfigure}[]{0.46\linewidth}
        \centering
        \adjincludegraphics[width=\linewidth,trim={{.00\width} 0 {.08\width} 0},clip]{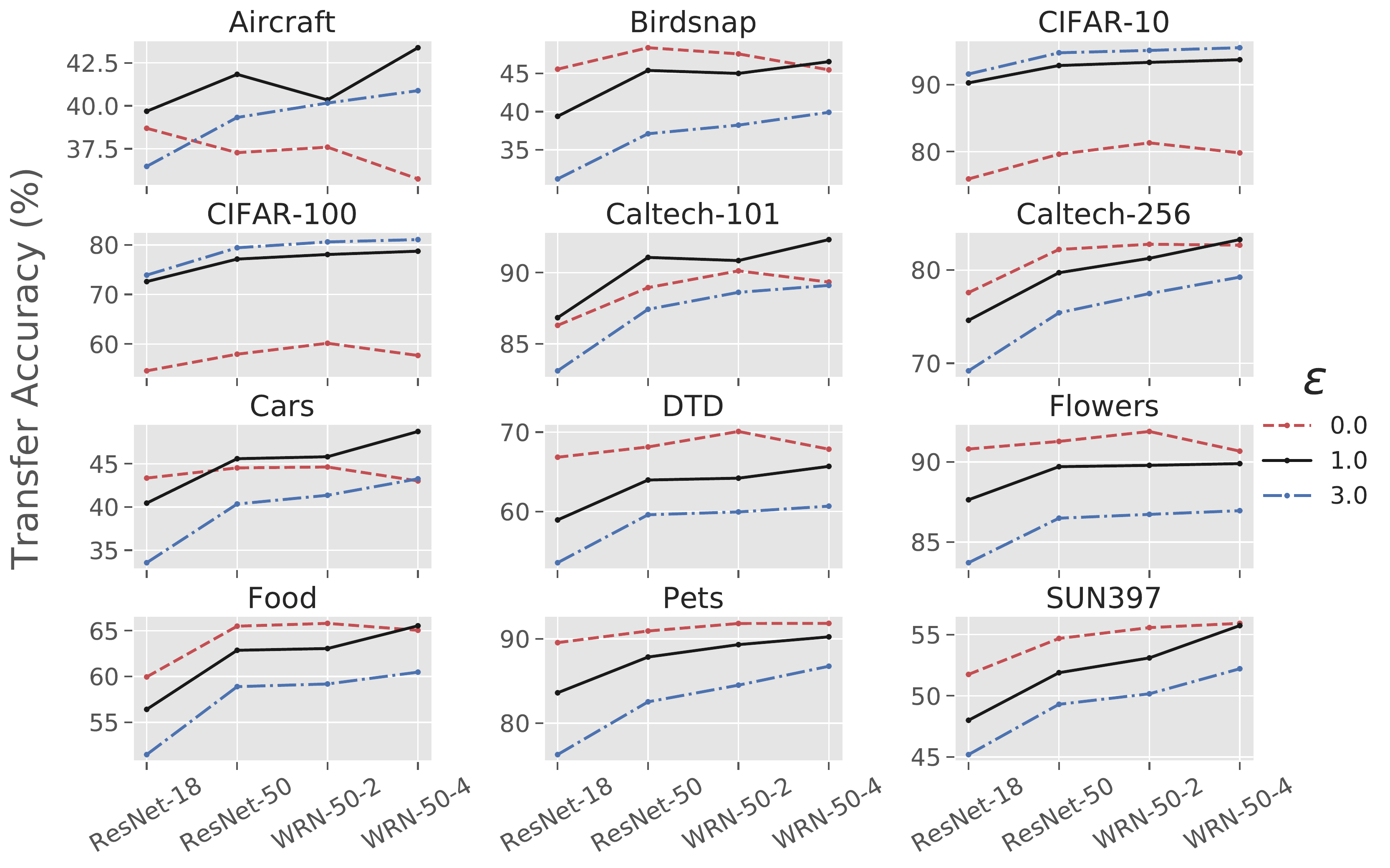}
        \caption{\textbf{Fixed-feature} transfer}
    \end{subfigure}
    \begin{subfigure}[]{0.49\linewidth}
        \centering
        \adjincludegraphics[width=\linewidth,trim={{.06\width} 0 {.0\width} 0},clip]{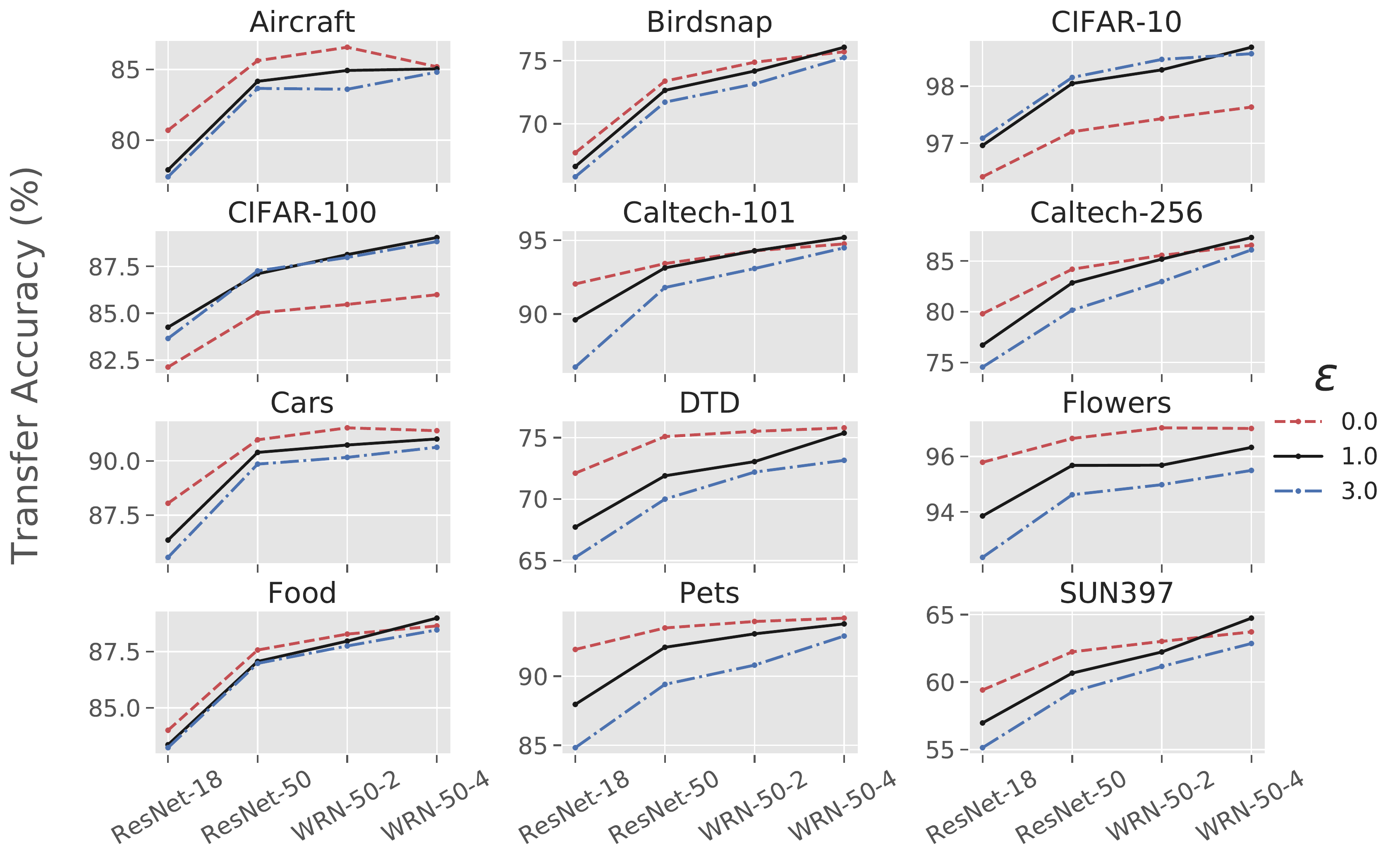}
        \caption{\textbf{Full-network} transfer}
    \end{subfigure}
    \caption{Varying width and model robustness while transfer learning from
      ImageNet to various datasets. Generally, as width increases, transfer
      learning accuracies of standard models generally plateau or level off
      while those of robust models steadily increase. More values of $\varepsilon$ are in Appendix~\ref{app:omitted-results}.}
    \label{fig:width-effect}
\end{figure}

Our experiments also reveal a contrast between robust and standard models in how
their transfer performance scales with model width. 
\citet{azizpour2015factors}, find that although increasing network depth
improves transfer performance, increasing width hurts it. 
Our results corroborate this trend for standard networks, but indicate that it
does {\em not} hold for robust networks, at least in the regime of widths tested.
Indeed, Figure~\ref{fig:width-effect} plots results for the three widths of
ResNet-50 studied here (x1, x2, and x4), along with a ResNet-18 for reference:
as width increases, transfer performance plateaus and decreases for standard
models, but continues to steadily grow for robust models.
This suggests that scaling network width may further increase the transfer
performance gain of robust networks over the standard ones. (This increase
comes, however, at a higher computational cost.)  

%% file: sections/effective_epsilon.tex
\subsection{Optimal robustness levels for downstream tasks}
\label{sec:effective-epsilon-hypothesis}
We observe that although the best robust models often outperform the best
standard models, the optimal choice of robustness parameter $\varepsilon$ varies
widely between datasets. For example, when transferring to CIFAR-10 and
CIFAR-100, the optimal $\varepsilon$ values were $3.0$ and $1.0$, respectively.
In contrast, smaller values of $\varepsilon$ (smaller by an order of magnitude) tend
to work better for the rest of the datasets.

One possible explanation for this variability in the optimal choice of
$\varepsilon$ might relate to dataset granularity.
We hypothesize that on datasets where leveraging finer-grained features are necessary (i.e.,
where there is less norm-separation between classes in the input space), the most
effective values of $\varepsilon$  will be much smaller than for a dataset where leveraging more coarse-grained features suffices. To illustrate this, consider a binary classification task consisting of image-label
pairs $(x, y)$, where the correct class for an image $y \in \{0, 1\}$ is determined
by a single pixel, i.e., $x_{0,0} = \delta \cdot y$, and $\ x_{i, j} = 0$, otherwise.
We would expect transferring a standard model onto this dataset to yield perfect
accuracy regardless of $\delta$, since the dataset is perfectly separable. On
the other hand, a robust model is trained to be invariant to perturbations of
norm $\varepsilon$---thus, if $\delta < \varepsilon$, the dataset will not
appear separable to the standard model and so we expect transfer to be
less successful. So, the smaller the $\delta$ (i.e., the larger the ``fine grained-ness'' of
the dataset), the smaller the $\varepsilon$ must be for successful transfer.

\paragraph{Unifying dataset scale.}
We now present evidence in support of our above hypothesis. Although we lack a
quantitative notion of granularity (in reality, features are not simply singular
pixels), we consider image resolution as a crude proxy. Since we scale target
datasets to match ImageNet dimensions, each pixel in a
low-resolution dataset (e.g., CIFAR-10) image
translates into several pixels in transfer, thus inflating datasets'
separability. Drawing from this observation, we attempt to calibrate the
granularities of the 12 image classification datasets used in this work, by
first downscaling all the images to the size of CIFAR-10 ($32\times 32$), and then
upscaling them to ImageNet size once more. We then repeat the fixed-feature
regression experiments from prior sections, plotting the
results in Figure~\ref{fig:downscaling-experiment} (similar results for full-network transfer are presented in Appendix~\ref{app:omitted-results}). After controlling for
original dataset dimension, the datasets' epsilon vs. transfer accuracy
curves all behave almost identically to CIFAR-10 and CIFAR-100 ones. Note that while
this experimental data supports our hypothesis, we do not take the evidence as an ultimate one and further exploration is needed to reach definitive conclusions.

\begin{figure}[t]
        \centering
        \includegraphics[width=\linewidth]{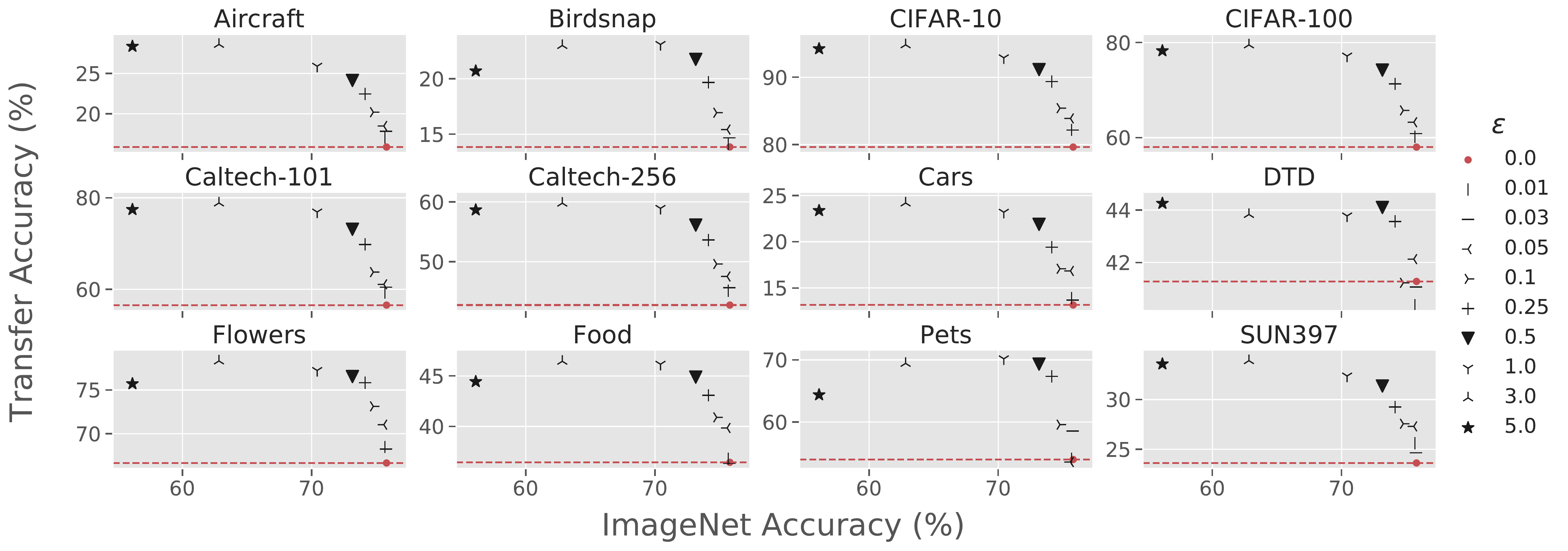}
    \caption{\textbf{Fixed-feature} transfer accuracies of various datasets that
      are down-scaled to $32\times 32$ before being up-scaled again to ImageNet
      scale and used for transfer learning. The accuracy curves are closely
      aligned, unlike those of
      Figure~\ref{fig:main-small-datasets-transfer-results-logistic-regression},
      which illustrates the same experiment without downscaling.}
    \label{fig:downscaling-experiment}

\end{figure}

%% file: sections/stylized_comparison.tex
\subsection{Comparing adversarial robustness to texture robustness}

\begin{figure}[!ht]
    \begin{subfigure}[]{0.4\linewidth}
        \adjincludegraphics[width=\linewidth,trim={0 {4.5em} 0 0},clip]{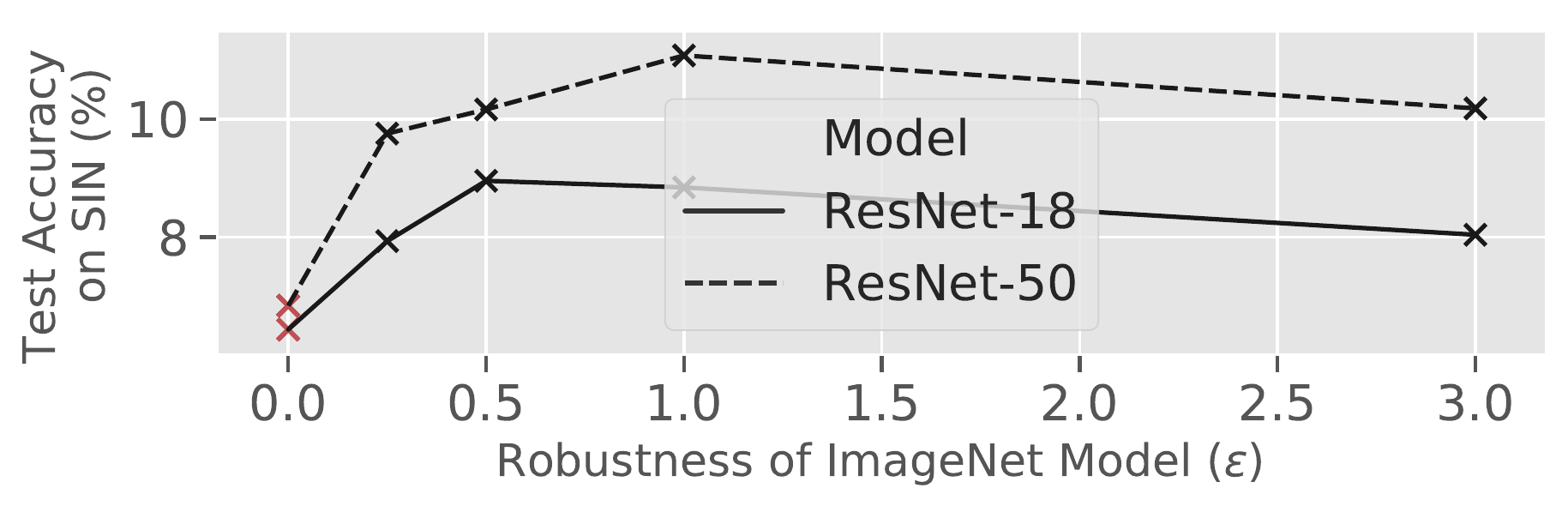}
        \includegraphics[width=\linewidth]{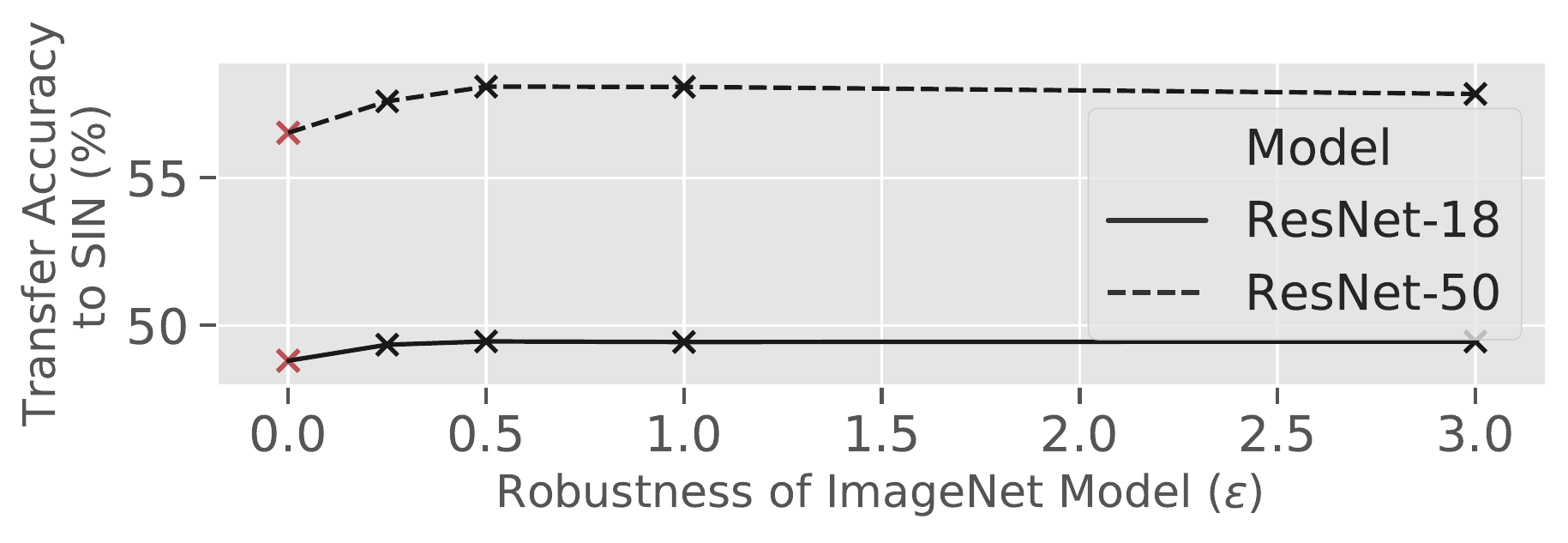}
        \caption{Stylized ImageNet Transfer}
        \label{fig:stylized-evaluate}
    \end{subfigure}
    \begin{subfigure}[]{0.59\linewidth}
        \centering
        \adjincludegraphics[height=2.9cm,trim={{0.00\width} 0 {0.01\width} 0},clip]{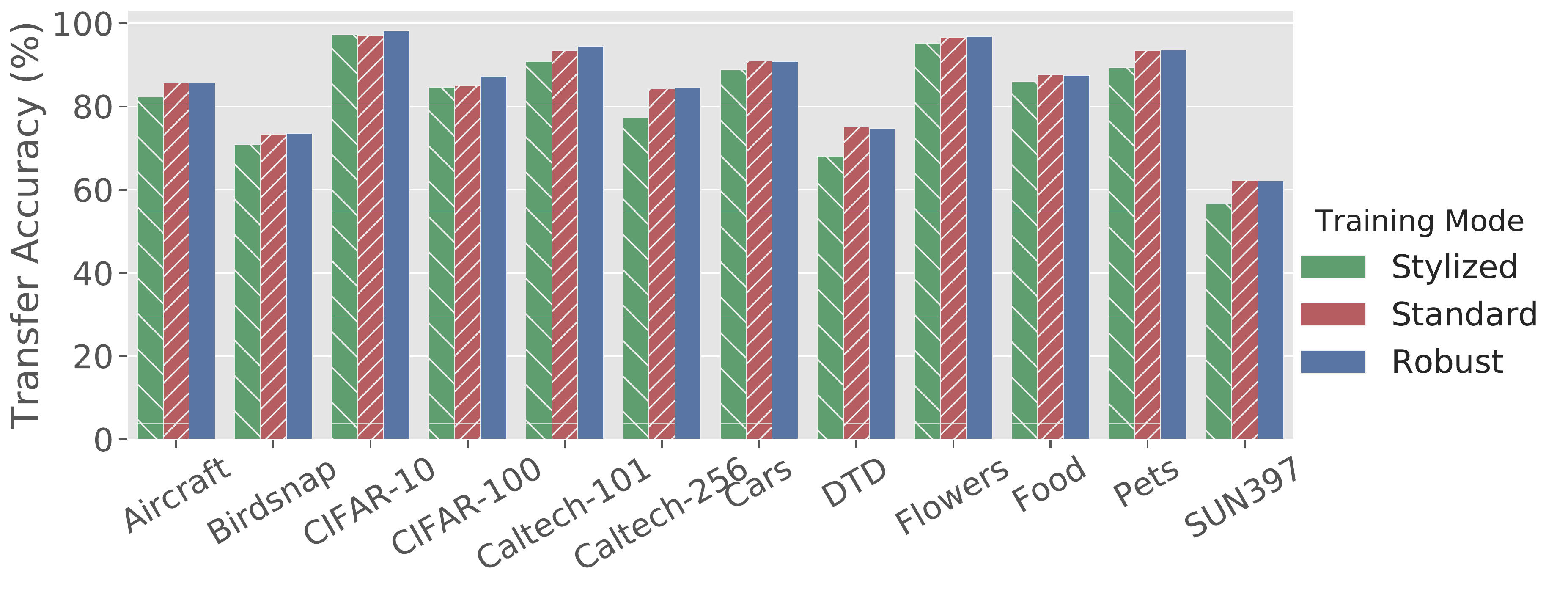}
        \caption{Transfer to Standard Datasets using a ResNet-50}
        \label{fig:stylized-transfer}
    \end{subfigure}
    \caption{We compare standard, stylized and robust ImageNet models on standard transfer tasks (and to stylized ImageNet).}
\end{figure}

We now investigate the effects of adversarial robustness on transfer learning
performance in comparison to other invariances commonly imposed on deep neural
networks.
Specifically, we consider texture-invariant~\citep{geirhos2018imagenettrained}
models, i.e., models trained on the Stylized ImageNet
(SIN)~\citep{geirhos2018imagenettrained} dataset. 
Figure~\ref{fig:stylized-transfer} 
shows that transfer learning from adversarially robust models outperforms
transfer learning from texture-invariant models on all considered datasets. 

Finally, we use the SIN dataset to further re-inforce the benefits conferred by
adversarial robustness. Figure~\ref{fig:stylized-evaluate} top shows that
robust models outperform standard imagenet models when evaluated (top) or
fine-tuned (bottom) on Stylized-ImageNet.

%% file: sections/related.tex
A number of works study transfer learning with CNNs \citep{donahue2014decaf,
  chatfield2014return, razavian2014cnn, azizpour2015factors}. Indeed, transfer
learning has been studied in varied domains including medical imaging
\citep{mormont2018comparison}, language modeling \citep{conneau2018senteval},
and various object detection and segmentation related
tasks~\citep{ren2015faster, dai2016r, huang2017speed, chen2017deeplab}. In terms
of methods, others~\citep{agrawal2014analyzing, chatfield2014return,
  girshick2014rich, yosinski2014transferable,
  azizpour2015factors,lin2015bilinear, huh2016makes, chu2016best} show that
fine-tuning typically outperforms frozen feature-based methods.
As discussed throughout this paper, several prior
works~\citep{azizpour2015factors, huh2016makes, kornblith2019better, 
zamir2018taskonomy, kolesnikov2019big, sun2017revisiting,
mahajan2018exploring, yosinski2014transferable} have investigated factors
improving or otherwise affecting transfer learning performance. Recently
proposed methods have achieved state-of-the-art performance on downstream tasks by
scaling up transfer learning techniques~\citep{huang2018gpipe,
kolesnikov2019big}. 

On the adversarial robustness front, many works---both
empirical (e.g., \citep{madry2018towards, miyato2018virtual, balaji2019instance,
zhang2019theoretically}) and  certified (e.g.,
\citep{lecuyer2018certified, weng2018towards, wong2018provable,
raghunathan2018certified, cohen2019certified, salman2019provably,
yang2020randomized})---significantly increase model resilience to adversarial 
examples~\citep{biggio2013evasion, szegedy2014intriguing}. 
A growing body of research has studied the {\em features} learned by these robust
networks and suggested that they improve upon those learned by standard networks
(cf.~\citep{ilyas2019adversarial, engstrom2019learning, santurkar2019image,
allenzhu2020feature, kim2019bridging, kaur2019perceptually} and references). On
the other hand, prior studies have also identified theoretical and empirical
tradeoffs between standard accuracy and adversarial
robustness~\citep{tsipras2019robustness, bubeck2018adversarial,
su2018robustness, raghunathan2019adversarial}. At the intersection of robustness
and transfer learning, \citet{shafahi2019adversarially} investigate transfer
learning for increasing downstream-task adversarial robustness (rather than
downstream accuracy, as in this work). \citet{aggarwal2020benefits} find
that adversarially trained models perform better at downstream zero-shot
learning tasks and weakly-supervised object localization. Finally, concurrent to our
work,~\citep{utrera2020adversarially} also study the transfer performance of 
adversarially robust networks. Our studies reach similar conclusions and are
otherwise complementary: here we study a larger set of downstream datasets and
tasks and analyze the effects of model accuracy, model width, and data
resolution;~\citet{utrera2020adversarially} study the effects of training
duration, dataset size, and also introduce an influence 
function-based analysis~\citep{koh2017understanding} to study the
representations of robust networks.  
For a detailed discussion of prior work, see Appendix~\ref{app:rel}.

%% file: sections/conclusion.tex
In this work, we propose using adversarially robust models for
transfer learning.
We compare transfer learning performance of robust and standard
models on a suite of 12 classification tasks, object
detection, and instance segmentation. 
We find that adversarial robust neural networks consistently match or improve upon the
performance of their standard counterparts, despite having lower ImageNet
accuracy. 
We also take a closer look at the behavior of adversarially robust networks, and
study the interplay between ImageNet 
accuracy, model width, robustness, and transfer performance.

%% file: sections/acknowledgements.tex
Work supported in part by the NSF awards CCF-1553428, CNS-1815221, the Open
Philanthropy Project AI Fellowship, and the Microsoft Corporation. This material is
based upon work supported by the Defense Advanced Research Projects Agency
(DARPA) under Contract No. HR001120C0015.

%% file: sections/app_experimental_setup.tex
\subsection{Pretrained ImageNet models}
\label{app:pretrained-models}
In this paper, we train a number of standard and robust ImageNet models on various architectures. 
These models are used for all the various transfer learning experiments.

\paragraph{Architectures}
We experiment with several standard architectures from the PyTorch's Torchvision\footnote{These models can be found here \url{https://pytorch.org/docs/stable/torchvision/models.html}}. 
These models are shown in Tables~\ref{table:l2-classification-networks}\&\ref{table:linf-classification-networks}.\footnote{WRN-50-2 and WRN-50-4 refer to Wide-ResNet-50, twice and four times as wide, respectively.}

\begin{table}[htbp]
    \centering
    \caption{The clean accuracies of standard and $\ell_2$-robust ImageNet classifiers used in our paper.}
    \label{table:l2-classification-networks}
        \begin{tabular}{@{}lcccccccccc@{}}
        \toprule
        & \multicolumn{10}{c}{\textbf{Clean ImageNet Top-1 Accuracy (\%)}} \\
        \midrule
        & \multicolumn{10}{c}{Robustness parameter $\varepsilon$} \\
        \cmidrule(lr){2-11}
        \textbf{Model}      & $0$  & $0.01$ & $0.03$ & $0.05$ & $0.1$ & $0.25$ & $0.5$ & $1$ & $3$ & $5$ \\ \midrule
        \textbf{ResNet-18} & 69.79 & 69.90 & 69.24 & 69.15 & 68.77 & 67.43 & 65.49 & 62.32 & 53.12 & 45.59 \\
        \textbf{ResNet-50} & 75.80 & 75.68 & 75.76 & 75.59 & 74.78 & 74.14 & 73.16 & 70.43 & 62.83 & 56.13 \\
        \textbf{WRN-50-2 } & 76.97 & 77.25 & 77.26 & 77.17 & 76.74 & 76.21 & 75.11 & 73.41 & 66.90 & 60.94 \\
        \textbf{WRN-50-4 } & 77.91 & 78.02 & 77.87 & 77.77 & 77.64 & 77.10 & 76.52 & 75.51 & 69.67 & 65.20 \\
        \bottomrule
    \end{tabular}
    \vskip 0.3cm
    \begin{tabular}{@{}lcccccc@{}}
        \toprule
        & \multicolumn{6}{c}{\textbf{Clean ImageNet Top-1 Accuracy (\%)}} \\
        \midrule 
        & \multicolumn{6}{c}{\textbf{Model Architecture}} \\
        \cmidrule(lr){2-7}
        &  \textbf{A} &  \textbf{B} &  \textbf{C} &  \textbf{D} &  \textbf{E} &  \textbf{F} \\
            &  \textbf{DenseNet-161} &\textbf{ResNeXt50} &    \textbf{VGG16-bn} &  \textbf{MobileNet-v2} &  \textbf{ShuffleNet }&  \textbf{MNASNET }\\
        \midrule
        \textbf{$\varepsilon = 0$  } 
        &         77.37 &     77.32 &        73.66 &      65.26 &       64.25 &     60.97 \\
        
        \textbf{$\varepsilon = 3$  } 
        &         66.12 &    65.92 &         56.78 &      50.05 &       42.87 &     41.03 \\
        \bottomrule
    \end{tabular}    
\end{table}

\begin{table}[htbp]
    \caption{The clean accuracies of $\ell_\infty$-robust ImageNet classifiers.}
    \label{table:linf-classification-networks}
    \begin{center}
        \begin{tabular}{@{}lcccccc@{}}
        \toprule
        & \multicolumn{5}{c}{\textbf{Clean ImageNet Top-1 Accuracy (\%)}} \\
        \midrule
        & \multicolumn{5}{c}{Robustness parameter $\varepsilon$} \\
        \cmidrule(lr){2-6}
        \textbf{Model}      & $\frac{0.5}{255}$  & $\frac{1}{255}$ & $\frac{2}{255}$ & $\frac{4}{255}$ & $\frac{8}{255}$ \\ \midrule
        \textbf{ResNet-18} & 66.13 & 63.46 & 59.63 & 52.49 & 42.11 \\
        \textbf{ResNet-50} & 73.73 & 72.05 & 69.10 & 63.86 & 54.53 \\
        \textbf{WRN-50-2 } & 75.82 & 74.65 & 72.35 & 68.41 & 60.82 \\
        \bottomrule
    \end{tabular}
\end{center}
\end{table}

\paragraph{Training details}
We fix the training procedure for all of these models. 
We train all the models from scratch using SGD with batch size of 512, momentum of $0.9$, and weight decay of $1e-4$.
We train for 90 epochs with an initial learning rate of $0.1$ that drops by a factor of $10$ every 30 epochs.

For \textbf{Standard Training}, we use the standard cross-entropy multi-class classification loss.
For \textbf{Robust Training}, we use adversarial training \citep{madry2018towards}. 
We train on adversarial examples generated within maximum allowed perturbations $\ell_2$ of $\varepsilon$ $ \in \{0.01,0.03,0.05,0.1,0.25,0.5,1,3,5 \}$ and $\ell_{\infty}$ perturbations of $\varepsilon$ $\in \{\frac{0.5}{255}, \frac{1}{255} ,\frac{2}{255} ,\frac{4}{255} ,\frac{8}{255}  \}$
using 3 attack steps and a  step size of $\frac{\varepsilon \times 2}{3}$.

\subsection{ImageNet transfer to classification datasets}
\label{app:transfer-to-small-datasets}

\subsubsection{Datasets}
\label{app:classification-datasets}
\begin{table}[!h]
\caption{Classification datasets used in this paper.}
\label{table:datasets}
    \begin{center}
        \begin{small}
        \begin{tabular}{@{}lccc@{}}
        \toprule
        \textbf{Dataset}           & \textbf{Classes} & \textbf{Size (Train/Test)} & Accuracy Metric\\ \midrule
        Birdsnap \citep{berg2014birdsnap}                  & 500     & 32,677/8,171  & Top-1      \\ 
        Caltech-101 \citep{fei2004learning} & 101 & 3,030/5,647 & Mean Per-Class\\
        Caltech-256 \citep{griffin2007caltech} & 257 & 15,420/15,187 & Mean Per-Class\\ 
        CIFAR-10 \citep{krizhevsky2009learning}                  & 10      & 50,000/10,000  & Top-1   \\ 
        CIFAR-100 \citep{krizhevsky2009learning}                 & 100     & 50,000/10,000  & Top-1     \\ 
        Describable Textures (DTD) \citep{cimpoi2014describing} & 47      & 3,760/1,880  & Top-1       \\ 
        FGVC Aircraft \citep{maji2013fine}             & 100     & 6,667/3,333  & Mean Per-Class       \\ 
        Food-101 \citep{bossard2014food}                  & 101     & 75,750/25,250  & Top-1     \\ 
        Oxford 102 Flowers \citep{nilsback2008automated}        & 102     & 2,040/6,149  & Mean Per-Class         \\
        Oxford-IIIT Pets \cite{parkhi2012cats} & 37 & 3,680/3,669 & Mean Per-Class\\
        SUN397 \citep{xiao2010sun}                    & 397     & 19,850/19,850  & Top-1     \\ 
        Stanford Cars \citep{krause2013collecting}              & 196     & 8,144/8,041  & Top-1      \\ 
        \bottomrule
    \end{tabular}
    \end{small}
    \end{center}
\end{table}

We test transfer learning starting from ImageNet pretrained models on classification datasets that are used in \citep{kornblith2019better}. 
These datasets vary in size the number of classes and datapoints. 
The details are shown in Table~\ref{table:datasets}.

\subsubsection{Fixed-feature Transfer}
\label{app:logistic-regression-params}
For this type of transfer learning, we \textit{freeze} the weights of the ImageNet pretrained model\footnote{For all of our experiments, we do not freeze the batch statistics, only its weights.}, and replace the last fully connected layer with a random initialized one that fits the transfer dataset.
We train only this new layer for 150 epochs using SGD with batch size of 64, momentum of 0.9, weight decay of $5e-4$, and an initial lr $\in \{0.01, 0.001\}$ that drops by a factor of 10 every 50 epochs.
We use the following standard data-augmentation methods:

\begin{lstlisting}
    TRAIN_TRANSFORMS = transforms.Compose([
        transforms.RandomResizedCrop(224),
        transforms.RandomHorizontalFlip(),
        transforms.ToTensor(),
    ])
    TEST_TRANSFORMS = transforms.Compose([
        transforms.Resize(256),
        transforms.CenterCrop(224),
        transforms.ToTensor()
    ])
\end{lstlisting}

\subsubsection{Full-network transfer}
\label{app:finetuning-params}
For full-network transfer learning, we use the exact same hyperparameters as the fixed-feature setting, but we \textit{do not freeze} the weights of the pretrained ImageNet model.

\subsection{Unifying dataset scale}
For this experiment, we follow the exact experimental setup of \ref{app:transfer-to-small-datasets} with the only modification being resizing all the datasets to $32\times 32$ then do dataugmentation as before:

\begin{lstlisting}
    TRAIN_TRANSFORMS = transforms.Compose([
        transforms.Resize(32),
        transforms.RandomResizedCrop(224),
        transforms.RandomHorizontalFlip(),
        transforms.ToTensor(),
    ])
    TEST_TRANSFORMS = transforms.Compose([
        transforms.Resize(32),
        transforms.Resize(256),
        transforms.CenterCrop(224),
        transforms.ToTensor()
    ])
\end{lstlisting}

\subsection{Replicate our results}
We desired simplicity and kept reproducibility in our minds when conducting our experiments, so we use standard hyperparameters and minimize the number of tricks needed to replicate our results. We open source all the standard and robust ImageNet models that we use in our paper, and our code is available at  \url{https://github.com/Microsoft/robust-models-transfer}.

\clearpage
\section{Transfer Learning with \texorpdfstring{$\ell_{\infty}$-}-robust ImageNet models}
We investigate how well other types of robust ImageNet models do in transfer learning.

\input{latex_tables/small_datasets_linf_models.tex}

\section{Object Detection and Instance Segmentation}
\label{app:det}
In this section we provide more experimental details, and results, relating to
our object detection and instance segmentation experiments.

\begin{figure}[h]
\centering
\includegraphics[width=.95\textwidth]{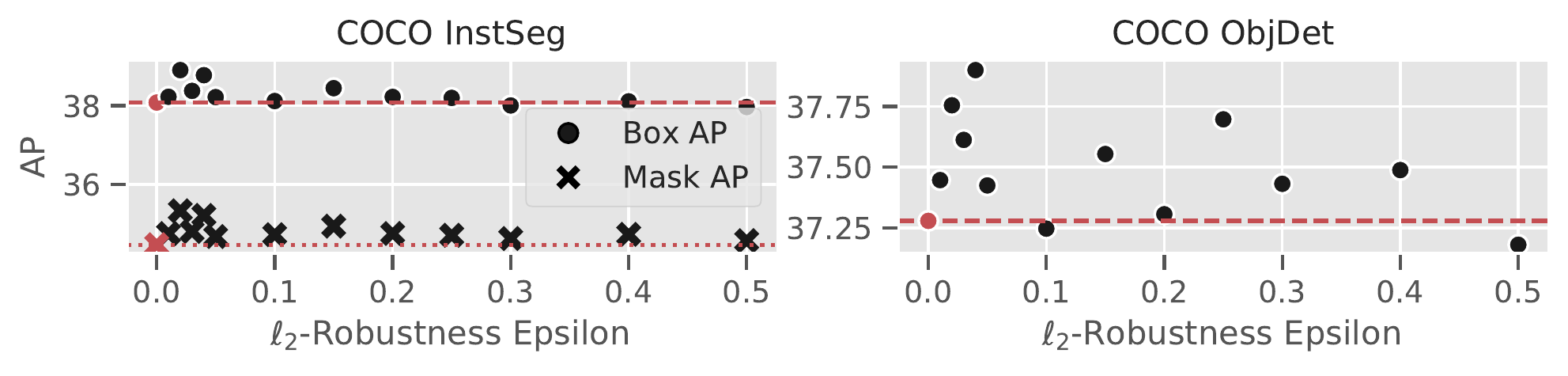}
\vspace{1em}
\small
 \caption{AP of instance segmentation and object detection models with backbones
   initialized with $\varepsilon$-robust models before training. Robust
   backbones generally lead to better AP, and the best robust backbone always
   outperforms the standard-trained backbone for every task.}
\label{app:eps_vs_ap_1x}
\end{figure}

\paragraph{Experimental setup.}
We use only standard configurations from Detectron2\footnote{See:
\url{https://github.com/facebookresearch/detectron2/blob/master/MODEL_ZOO.md}
For all COCO tasks we used ``R50-FPN'' configurations (1x and 3x, described
further in this section), and for VOC we used the ``R50-C4'' configuration.} to
train models. For COCO tasks, compute limitations made training from every
$\varepsilon$ initialization impossible. Instead, we trained from every
$\varepsilon$ initialization using a reduced learning rate schedule (the
corresponding 1x learning rate schedule in Detectron2) before training from
the top three $\varepsilon$ initializations (by Box AP) along with the standard
model using the full learning rate training schedule (the 3x schedule). Our
results for the 1x learning rate search are in
Figure~\ref{app:eps_vs_ap_1x}; our results, similar to those in
Section~\ref{sec:fullnetwork}, show that training from a robustly trained backbone
yields greater AP than training from a standard-trained backbone.

\paragraph{Baselines.}
We use standard ResNet-50 models from the torchvision
package\footnote{\url{https://pytorch.org/docs/stable/torchvision/index.html}}
using the Robustness library~\citep{robustness}. Detectron2 models were
originally trained for (and their configurations are tuned for) ResNet-50 models
from the original ResNet code
release\footnote{\url{https://github.com/KaimingHe/deep-residual-networks}},
which are slightly different from the torchvision ResNet-50s we use. It has been
previously noted that models trained from torchvision perform worse with
Detectron2 than these original models\footnote{See for both previous note and
model differences:
\url{https://github.com/facebookresearch/detectron2/blob/master/tools/convert-torchvision-to-d2.py}}.
Despite this, the best torchvision ResNet-50 models we train from robust
initializations dominate (without any additional hyperparameter searching) the
original baselines except for the COCO Object Detection task in terms of AP, in
which the original baseline has 0.07 larger Box AP\footnote{Baselines found
here:
\url{https://github.com/facebookresearch/detectron2/blob/master/MODEL_ZOO.md}}.

%% file: latex_tables/small_datasets_linf_models.tex
\begin{table}[!htbp]
\centering
\caption{Transfer Accuracy of standard vs $\ell_{\infty}$-robust ImageNet models on CIFAR-10 and CIFAR-100.}
\label{app:small-datasets-linf}
\begin{tabular}{lllrrrrrr}
\toprule
&& {} & \multicolumn{6}{c}{\textbf{Transfer Accuracy (\%)}} \\ \midrule
&& {} & \multicolumn{6}{c}{Robustness parameter $\varepsilon$} \\ \cmidrule(lr){4-9}
 & & & 0.0 & $\frac{0.5}{255}$ &   $\frac{1.0}{255}$ &   $\frac{2.0}{255}$ &   $\frac{4.0}{255}$ &   $\frac{8.0}{255}$ \\
\textbf{Dataset} & \textbf{Transfer Type} & \textbf{Model} &                   &       &       &       &       &       \\
\midrule
\multirow{4}{*}{\textbf{CIFAR-10}} & \multirow{2}{*}{\textbf{Full-network}} & \textbf{ResNet-18} &             96.05 & 96.85 & 96.80 & 96.98 & \textbf{97.04} & 96.79 \\
          &                     & \textbf{ResNet-50} &             97.14 & 97.69 & 97.84 & 97.98 & 97.92 & 98.01 \\
\cmidrule(lr){2-9}
          & \multirow{2}{*}{\textbf{Fixed-feature}} & \textbf{ResNet-18} &             75.02 & 87.13 & 89.01 & 89.07 & \textbf{90.56} & 89.18 \\
          &                     & \textbf{ResNet-50} &             78.16 & 90.55 & 91.51 & 92.74 & 93.35 & 93.68 \\
\cmidrule(lr){1-9}
\cmidrule(lr){2-9}
\multirow{4}{*}{\textbf{CIFAR-100}} & \multirow{2}{*}{\textbf{Full-network}} & \textbf{ResNet-18} &             81.70 & 83.66 & 83.46 & \textbf{83.98} & 83.55 & 82.82 \\
          &                     & \textbf{ResNet-50} &             84.75 & 86.12 & 86.48 & 87.06 & 86.90 & 86.21 \\
\cmidrule(lr){2-9}
          & \multirow{2}{*}{\textbf{Fixed-feature}} & \textbf{ResNet-18} &             53.86 & 68.52 & 70.83 & 72.00 & \textbf{72.19} & 69.78 \\
          &                     & \textbf{ResNet-50} &             55.57 & 72.89 & 74.16 & 76.22 & 77.17 & 76.70 \\
\bottomrule
\end{tabular}
\end{table}

%% file: sections/app_rel.tex
\section{Related Work}
\label{app:rel}

\paragraph{Transfer learning.}
Transfer learning has been investigated in a number of early works which extracted features from ImageNet CNNs and trained SVMs or logistic regression classifiers using these features on new datasets/tasks \citep{donahue2014decaf, chatfield2014return, razavian2014cnn}. These ImageNet features were shown to outperform hand-crafted features even on tasks different from ImageNet classification.\citep{razavian2014cnn, donahue2014decaf}. 
Later on, \citep{azizpour2015factors} demonstrated that transfer using deep networks is more effective than using wide networks across many transfer tasks.
A number of works has furthermore studied the transfer problem in the domain of medical imaging \citep{mormont2018comparison} and language modeling \citep{conneau2018senteval}. 
Besides, many of research in the literature has indicated that, specifically in computer vision, fine-tuning typically performs better than than classification based on freezed features \citep{agrawal2014analyzing, chatfield2014return, girshick2014rich, yosinski2014transferable, azizpour2015factors,lin2015bilinear, huh2016makes, chu2016best}. 

ImageNet pretrained networks have also been widely used as backbone models for various object detections models including Faster R-CNN and R-FCN \citep{ren2015faster, dai2016r}. More accurate ImageNet models tend to lead to  better overall object detection accuracy \citep{huang2017speed}. 
Similar usage is also common in image segmentation \citep{chen2017deeplab}.

Several works have studied how modifying the source dataset can affect the transfer accuracy. 
\citep{azizpour2015factors,huh2016makes} investigated the importance of the number of classes vs. number of images per class for learning better fixed image features, and these works have reached to conflicting conclusions \citep{kornblith2019better}. 
\citep{yosinski2014transferable} showed that freezing only the first layer of AlexNet does not affect the transfer performance between natural and manmade subsets of ImageNet as opposed to freezing more layers.  
Other works  demonstrated that transfer learning works even when the target dataset is large by transferring features learnt on a very large image datasets to ImageNet \citep{sun2017revisiting,mahajan2018exploring}.

More recently, \citep{zamir2018taskonomy} proposed a method to improve the efficiency of transfer learning  when labeled data from multiple domain are available. Furthermore, studied whether better ImageNet models transfer better to other datasets or not \citep{kornblith2019better}. 
It shows a strong correlation between the transfer accuracy of a pretrained ImageNet model (both for the logistic regression and finetuning settings) and the top-1 accuracy of these models on ImageNet. 
Finally, \citep{kolesnikov2019big} explored pretraining using enormous amount of data of around 300 million noisily labelled images, and showed improvements in transfer learning over pretraining on ImageNet for several tasks.

\paragraph{Transfer learning and robustness.}
A recent work \citep{shafahi2019adversarially} investigated the problem of
adversarially robust transfer learning: transferring adversarially robust
representations to new datasets while maintaining robustness on the downstream
task. While this work might look very similar to ours, there are two key
differences. The first is that this work investigates using robust source models
for the purpose of  improving/maintaining robustness on the downstream task, did
not investigate whether robust source models can improve the clean accuracy on
the downstream tasks. The second is that they point out that starting from a
standard trained ImageNet model leads to better natural accuracy when used for
downstream tasks, the opposite of what we show in the paper: we show that one
can get better transfer accuracies using robust, but less accurate, ImageNet
pretrained models. 

\paragraph{Robustness as a prior for learning representation.}
A major goal in deep learning is to learn robust high-level feature
representations of input data. However, current standard neural networks seem
to learn non-robust features that can be easily exploited to generate
adversarial examples. On the other hand, a number of recent papers have
argued that the features learned by adversarially robust models are less
vulnerable to adversarial examples, and at the same time are more
perceptually aligned with humans
\citep{ilyas2019adversarial,engstrom2019learning}. Specifically,
\citep{ilyas2019adversarial} presented a framework to study and disentangle
robust and non-robust features for standard trained networks. Concurrently,
\citep{engstrom2019learning} utilized this framework to show that robust
optimization can be re-cast as a tool for enforcing priors on the features
learned by deep neural networks. They showed that the representations learned
by robust models make significant progress towards learning a high-level
encoding of inputs.

%% file: sections/app-background.tex
\section{Background on Adversarially Robust Models}
\label{app:background}

\begin{figure}[ht]
    \centering
    \includegraphics[width=0.6\textwidth]{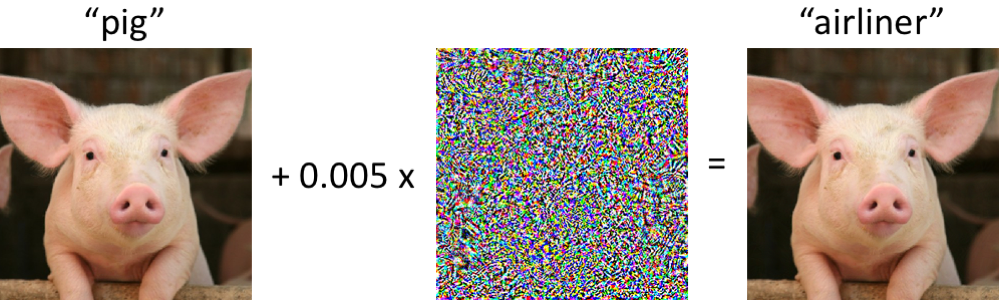} 
    \caption{An example of an adversarial attack: adding the imperceptible
    perturbation (middle) to a correctly classified pig (left) results in a
    near-identical image that is classified as ``airliner'' by an Inception-v3
    ImageNet model.} 
    \label{fig:adv_ex_ex}
\end{figure}

\paragraph{Adversarial examples in computer vision.} Adversarial
examples~\citep{biggio2013evasion,szegedy2014intriguing} (also referred to as
{\em adversarial attacks}) are imperceptible perturbations to natural inputs
that induce misbehaviour from machine learning---in this context
computer vision---systems. An illustration of such an attack is shown in
Figure~\ref{fig:adv_ex_ex}. The discovery of adversarial examples was a major
contributor to the rise of {\em deep learning security}, where prior work has
focused on both robustifying models against such
attacks (cf. \citep{goodfellow2015explaining,madry2018towards,
wong2018provable, raghunathan2018certified, cohen2019certified} and their
references), as well as testing the robustness of 
machine learning systems in ``real-world'' settings
(cf. \citep{papernot2017practical, athalye2018synthesizing, ilyas2018blackbox,
li2019adversarial, evtimov2018robust} and their references). A model that is
resilient to such adversarial examples is referred to as ``adversarially
robust.'' 

\paragraph{Robust optimization and adversarial training.} One of the canonical
methods for training an adversarially robust model is robust optimization.
Typically, we train deep learning models using empirical risk minimization (ERM) over
the training data---that is, we solve:
$$\min_\theta \frac{1}{n} \sum_{i=1}^{n} \mathcal{L}(x_i,y_i;\theta),$$
where $\theta$ represents the model parameters, $\mathcal{L}$ is a
task-dependent loss function (e.g., cross-entropy loss for classification), and
$\{(x_i, y_i) \sim \mathcal{D}\}$ are training image-label pairs.
In robust optimization (dating back to the work of~\citet{wald1945statistical}),
we replace this standard ERM objective with a {\em
robust} risk minimization objective:
$$\min_\theta \frac{1}{n} \sum_{i=1}^{n} \max_{x'; d(x_i,x') < \varepsilon}
\mathcal{L}(x', y_i),$$
where $d$ is a fixed but arbitrary norm. (In practice, $d$ is often assumed to be
an $\ell_p$ norm for $p \in \{2,\infty\}$---for the majority of this work we set
$p = 2$, so $d(x,x')$ is the Euclidean norm.)
In short, rather than minimizing the loss on only the training points, we
instead minimize the worst-case loss over a ball around each training point. 
Assuming the robust objective generalizes, it ensures that an
adversary cannot perturb a given test point $(x, y) \sim \mathcal{D}$ and
drastically increase the loss of the model. The
parameter $\varepsilon$ governs the desired robustness of the
model: $\varepsilon = 0$ corresponds to standard (ERM) training, and
increasing $\varepsilon$ results in models that are stable within larger and
larger radii. 

At first glance, it is unclear how to effectively solve the robust risk
minimization problem posed above---typically we use SGD to minimize risk, but 
here the loss function has an embedded maximization, so the corresponding SGD
update rule would be:
$$\theta_t \gets \theta_{t-1} - \eta \cdot \nabla_\theta \left(\max_{x'; d(x', x_i) <
\varepsilon} \mathcal{L}(x', y_i; \theta)\right).$$ 
Thus, to actually train an adversarially robust neural network,
\citet{madry2018towards} turn to inspiration from robust convex
optimization, where Danskin's theorem~\citep{danskin1967theory} says that for a
function $f(\alpha, \beta)$ that is convex in $\alpha$, 
$$\nabla_\alpha \left(\max_{\beta \in B} f(\alpha, \beta)\right) = 
\nabla_\alpha f(\alpha, \beta^*), \qquad\text{ where } \beta^* = \arg\max_\beta
f(\alpha, \beta) \text{ and $B$ is compact}.$$
Danskin's theorem thus allows us to write the gradient of a minimax problem in
terms of only the gradient of the inner objective, evaluated at its maximal
point. Carrying this intuition over to the neural network setting (despite the
lack of convexity) results in the popular {\em adversarial training}
algorithm~\citep{goodfellow2015explaining,madry2018towards}, where at each
training iteration, worst-case (adversarial)
inputs are passed to the neural network rather than standard unmodified inputs. Despite its simplicity,
adversarial training remains a competitive baseline for training adversarially
robust networks~\citep{rice2020overfitting}. Furthermore, recent works have
provided theoretical evidence for the success of adversarial training directly
in the neural network setting~\citep{gao2019convergence,allenzhu2020feature,
zhang2020overparameterized}. 

%% file: sections/app-omitted-figures.tex
\section{Omitted Figures}
\label{app:omitted-results}

\subsection{Full-network Transfer: additional results to Figure~\ref{fig:main-small-datasets-transfer-results-logistic-regression}}

\begin{figure}[!htbp]
    \centering
    \begin{subfigure}[]{0.7\linewidth}
        \adjincludegraphics[width=\linewidth]{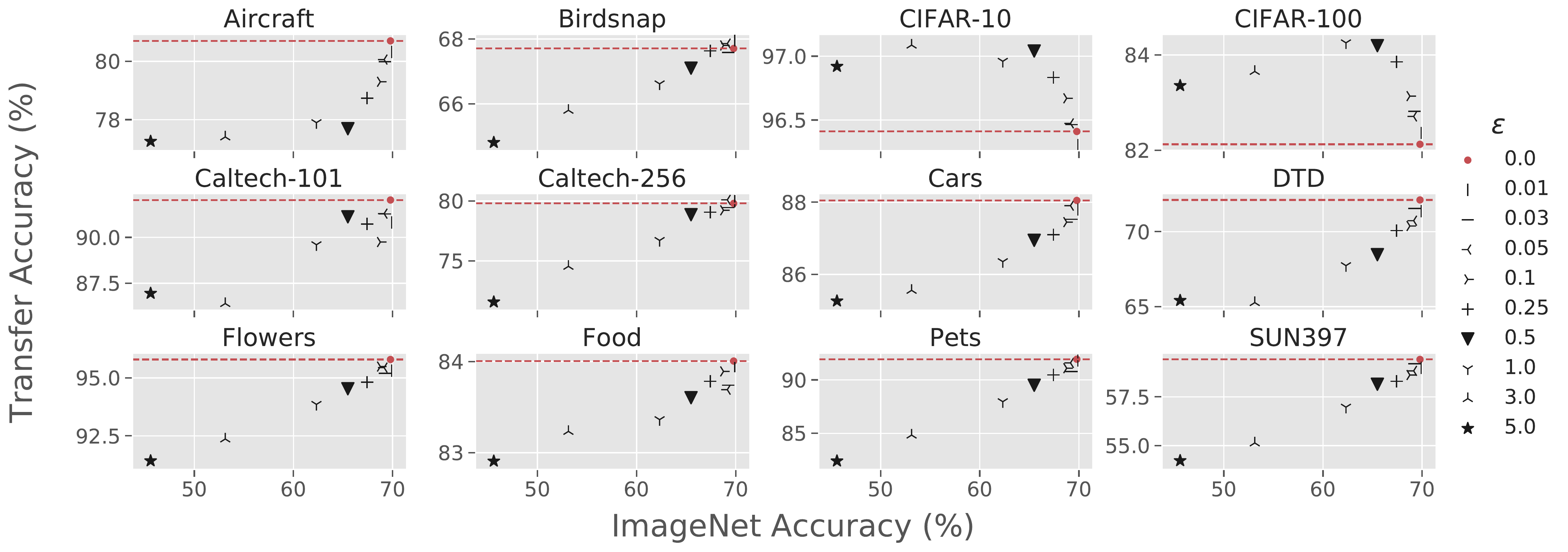}
        \caption{ResNet-18}
    \end{subfigure}
    \begin{subfigure}[]{0.7\linewidth}
        \adjincludegraphics[width=\linewidth]{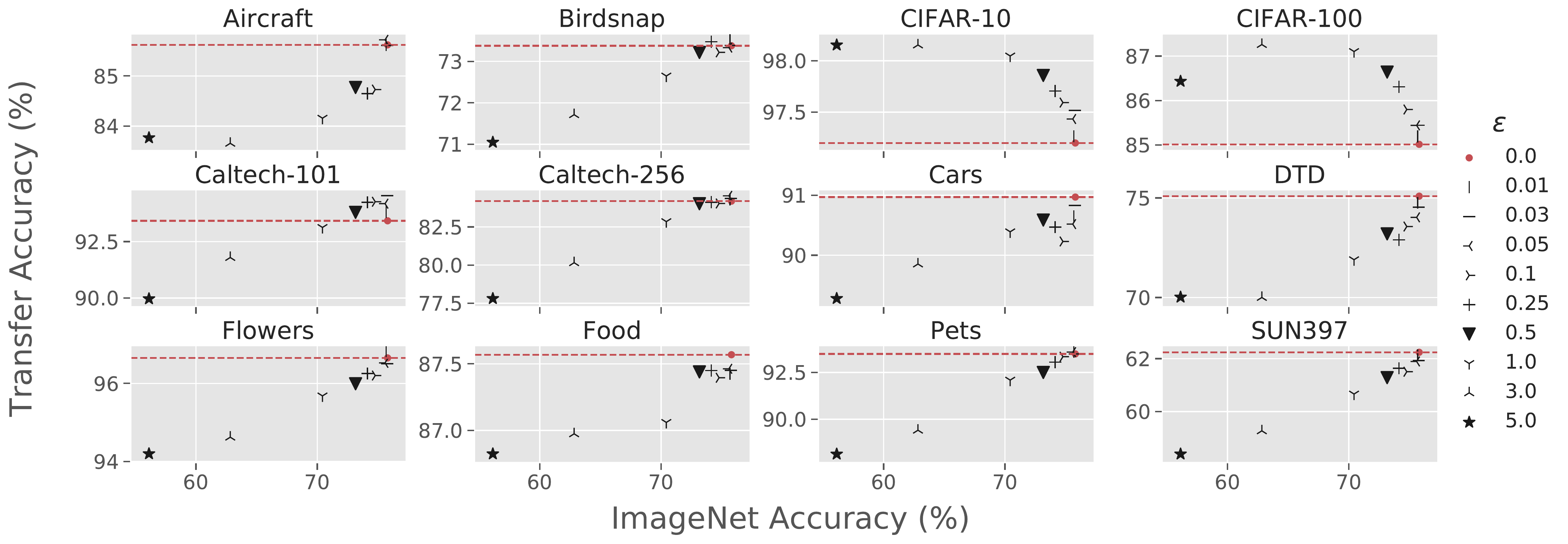}
        \caption{ResNet-50}
    \end{subfigure}
    \begin{subfigure}[]{0.7\linewidth}
        \adjincludegraphics[width=\linewidth]{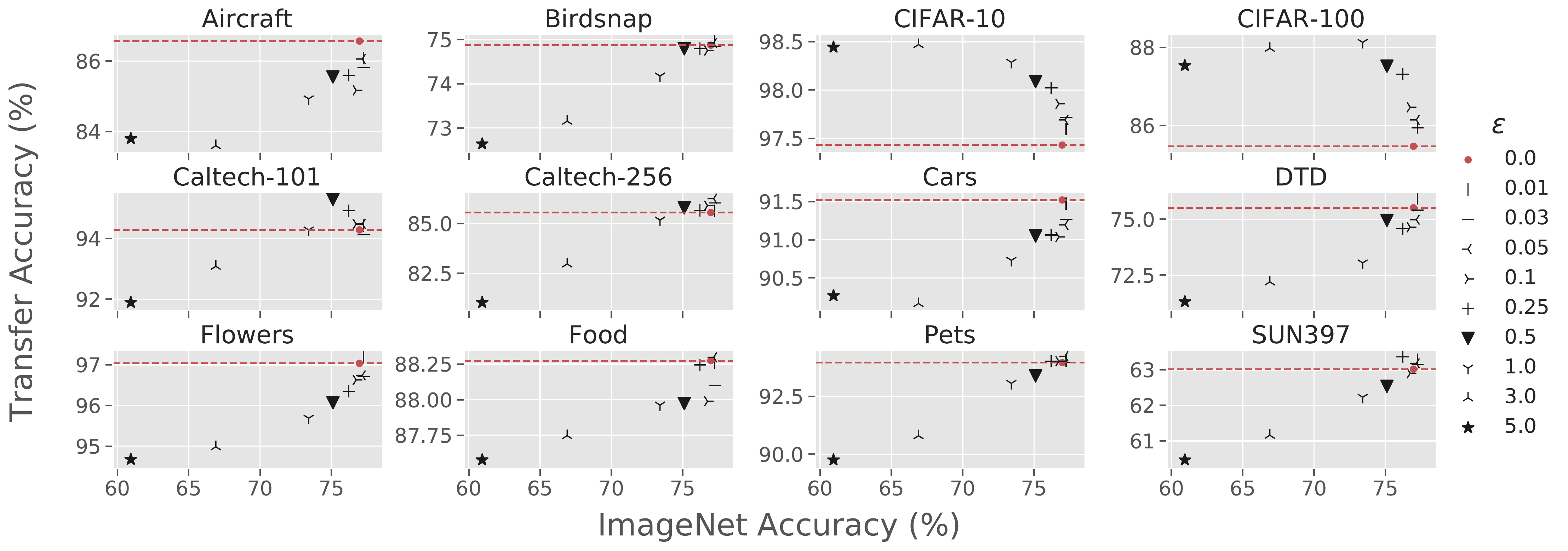}
        \caption{WRN-50-2}
    \end{subfigure}
    \begin{subfigure}[]{0.7\linewidth}
        \adjincludegraphics[width=\linewidth]{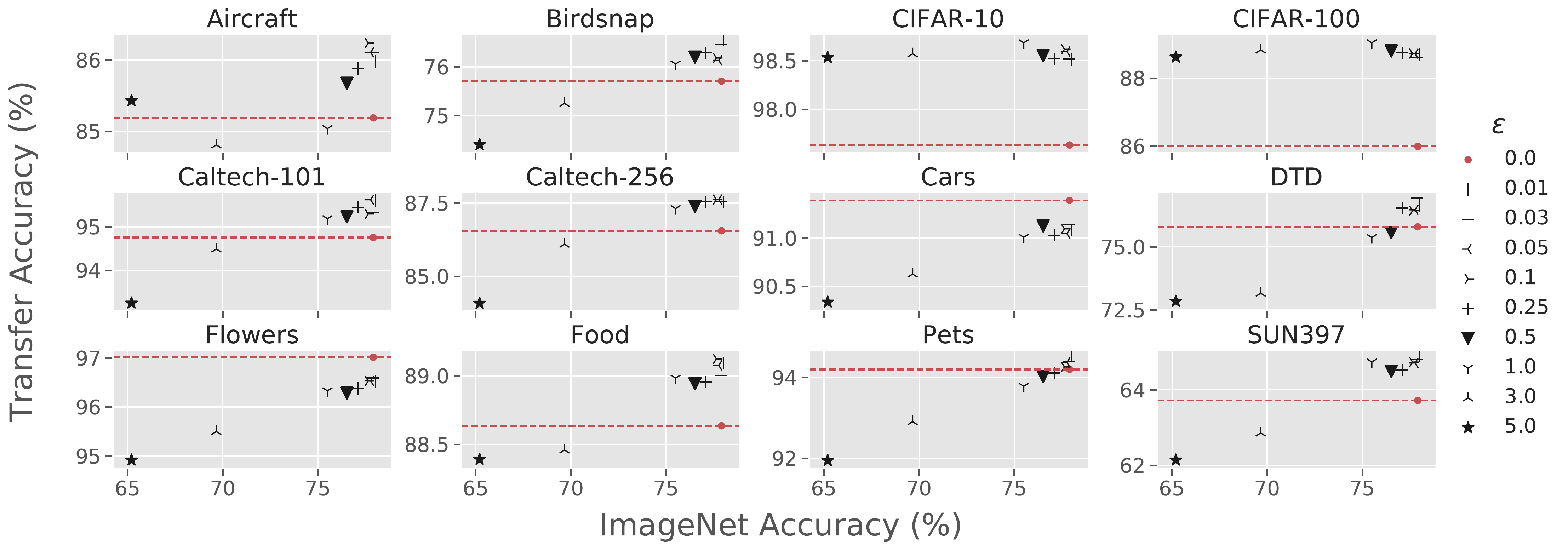}
        \caption{WRN-50-4}
    \end{subfigure}
 \caption{\textbf{Full-network} transfer accuracies of standard and robust
   ImageNet models to various image classification datasets. The linear
   relationship between accuracy and transfer performance does not hold;
   instead, for fixed accuracy, generally increased robustness yields higher
   transfer accuracy.}
 \label{fig:app-main-small-datasets-transfer-results-finetuning}
\end{figure}

\clearpage
\subsection{Varying architecture: additional results to Table~\ref{tab:linear_relation}}
\begin{table}[!htbp]
    \centering
    \caption{Source (ImageNet) and target accuracies, fixing robustness
    ($\varepsilon$) but varying architecture. When robustness is controlled for,
    ImageNet accuracy is highly predictive of (full-network) transfer performance.}
    \label{tab:linear_relation_appendix}
    \begin{tabular}{@{}llcccccc@{}cc@{}}
        \toprule
        & & \multicolumn{6}{c}{\textbf{Architecture} (see details in Appendix~\ref{app:pretrained-models})} & & \\
        \cmidrule(lr){3-9}
        \textbf{Robustness} & \textbf{Dataset} &  A &  B &  C &  D &  E &  F & &
        $R^2$ \\
        \midrule
        \multirow{2}{*}{Std ($\varepsilon=0$)  } & ImageNet &    77.37 & 77.32 &
        73.66 & 65.26 & 64.25 & 60.97 && --- \\ 
        & CIFAR-10 &         97.84 &  97.47  &        96.08 &     95.86 &      95.82
        &    95.55 & & 0.79 \\
        & CIFAR-100 & 86.53 &  85.53  &        82.07 &     80.02 &      80.76 &    80.41 & & 0.82\\
        & Caltech-101 & 94.78 &   94.63 &        91.32 &     88.91 &      87.13 &    83.28 && 0.94\\
        & Caltech-256 & 86.22 &   86.33 &        82.23 &     76.51 &      75.81 &    74.90 &&0.98\\
        & Cars & 91.28 &   91.27 &        90.97 &     88.31 &      85.81 &    84.54 && 0.91 \\
        & Flowers & 97.93 &   97.29 &        96.80 &     96.25 &      95.40 &    72.06 &&0.44 \\
        & Pets & 94.55 &   94.26 &        92.63 &     89.78 &      88.59 &    82.69 && 0.87 \\
        \midrule
        \multirow{2}{*}{Adv ($\varepsilon=3$)  } & ImageNet &        66.12 & 65.92 &
        56.78 & 50.05 & 42.87 & 41.03 && --- \\
        & CIFAR-10 &  98.67 & 98.22 &   97.27 &  96.91 & 96.23 &  95.99 && 0.97 \\
        & CIFAR-100 &  88.65 &   88.32 &        84.14 &     83.32 &      80.92 &    80.52 &&0.97\\
        & Caltech-101 & 93.84 &   93.31 &        89.93 &     89.02 &      83.29 &    75.52 && 0.83\\
        & Caltech-256 & 84.35 &  83.05  &        78.19 &     74.08 &      69.19 &    70.04 &&0.99\\
        & Cars & 90.91 &  90.08  &        89.67 &     88.02 &      83.57 &    78.76 &&0.79\\
        & Flowers & 95.77 & 96.01   &        93.88 &     94.25 &      91.47 &    26.98 &&0.38\\
        & Pets & 91.85 &  91.46  &        88.06 &     85.63 &      80.92 &    64.90 &&0.72\\
        \bottomrule
    \end{tabular}    
    \end{table}
\subsection{Stylized ImageNet Transfer: additional results to Figure~\ref{fig:stylized-transfer}}
\begin{figure}[!htbp]
    \begin{subfigure}[]{0.5\linewidth}
        \centering
        \adjincludegraphics[height=2.9cm,trim={{0.005\width} 0 {0.2\width} 0},clip]{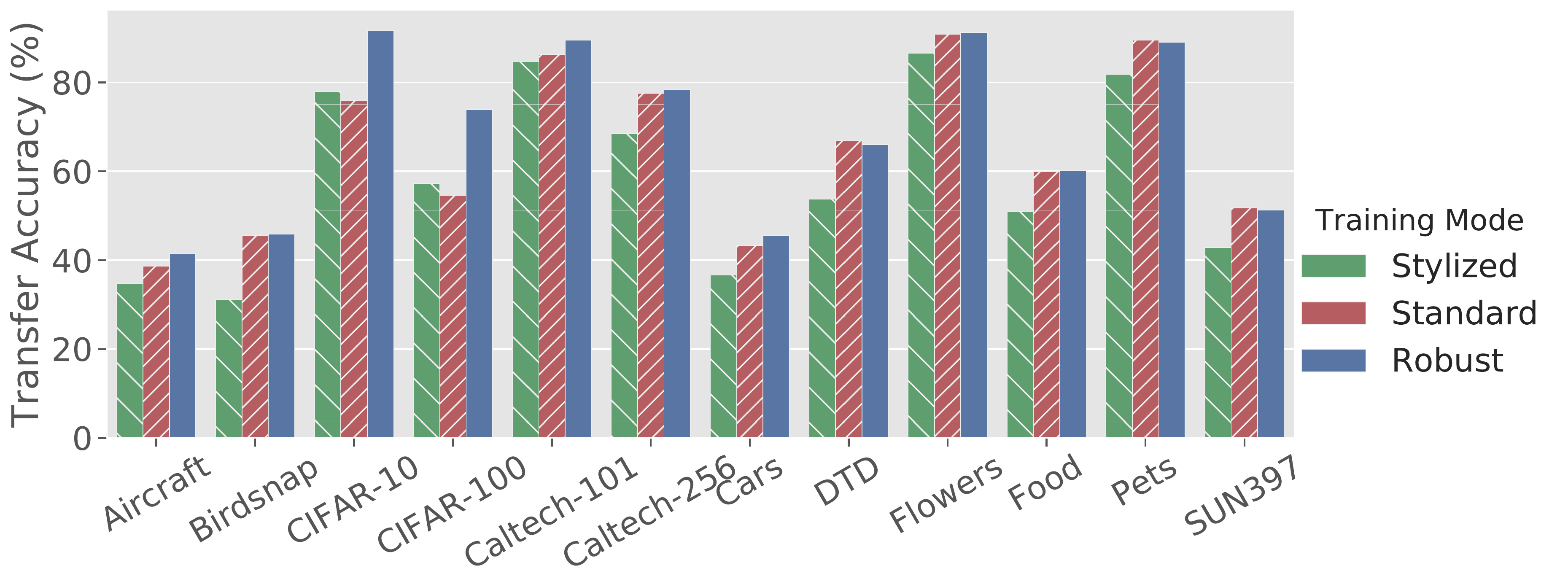}
        \caption{\textbf{Fixed-feature} ResNet-18}
    \end{subfigure}
    \begin{subfigure}[]{0.5\linewidth}
        \adjincludegraphics[height=2.9cm,trim={{0.035\width} 0 {0.01\width} 0},clip]{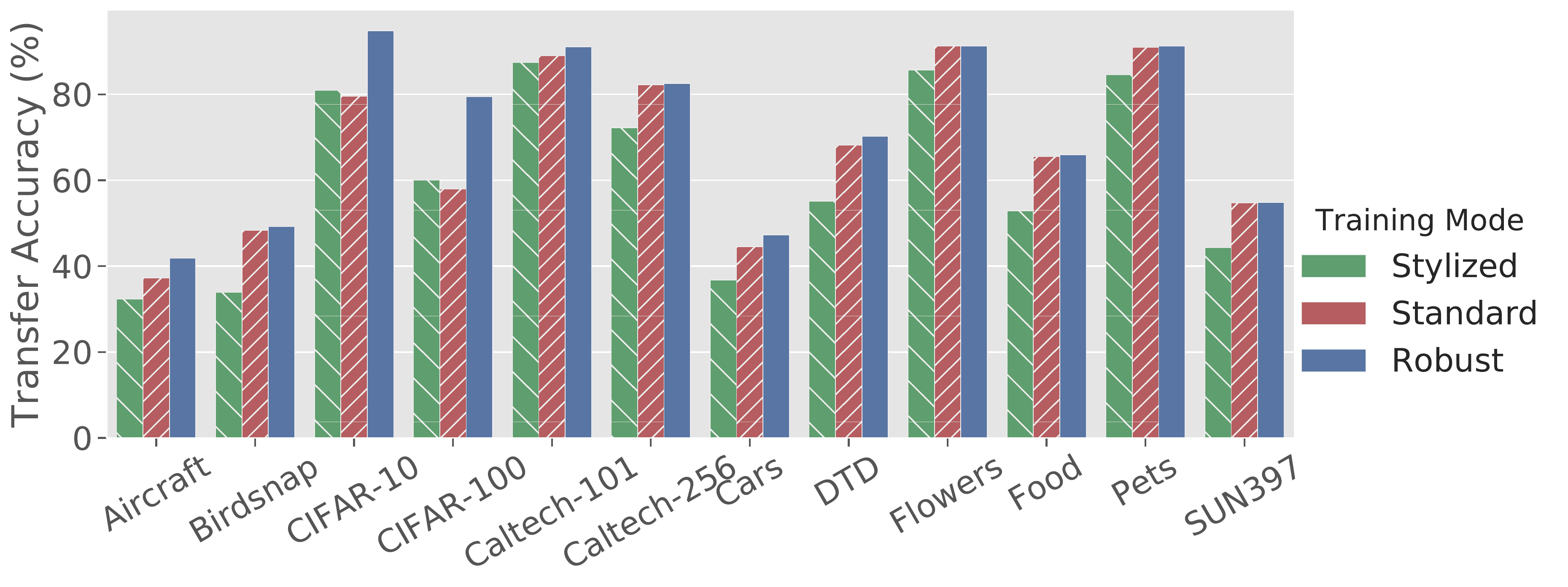}
        \caption{\textbf{Fixed-feature} ResNet-50}
    \end{subfigure}

    \begin{subfigure}[]{0.5\linewidth}
        \centering
        \adjincludegraphics[height=2.9cm,trim={{0.005\width} 0 {0.2\width} 0},clip]{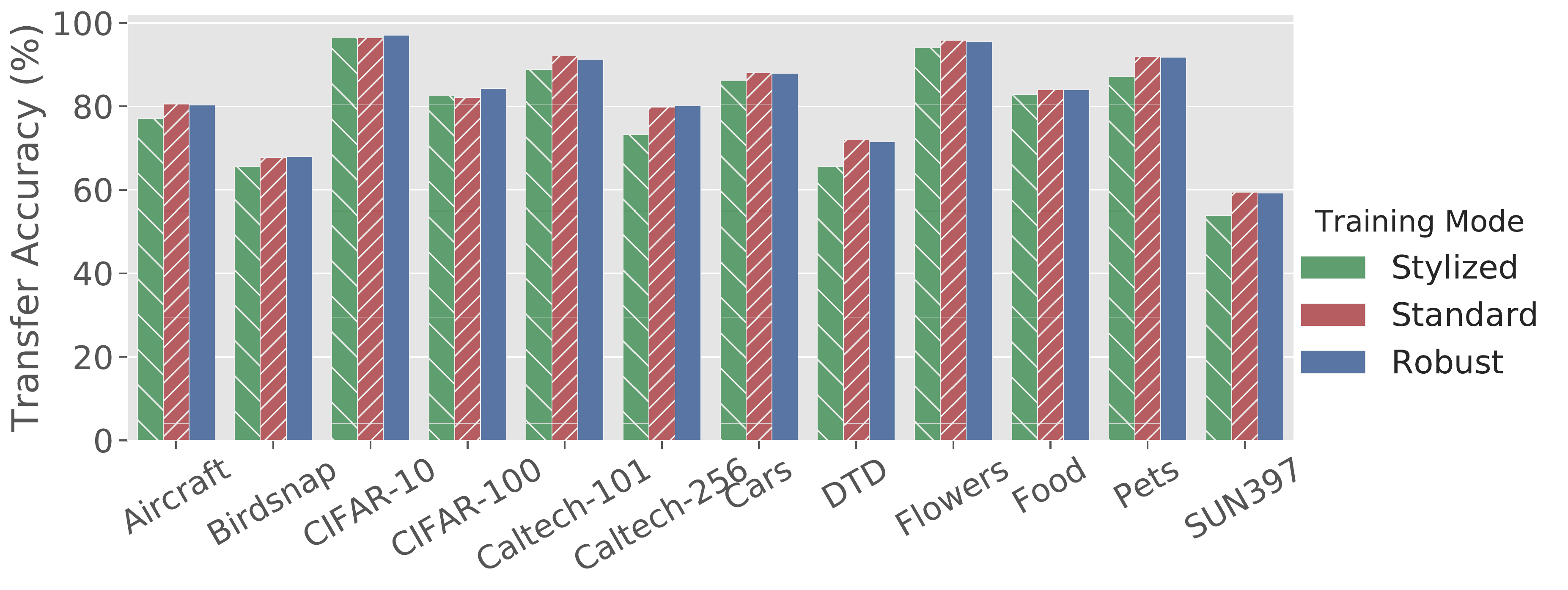}
        \caption{\textbf{Full-network} ResNet-18}
    \end{subfigure}
    \begin{subfigure}[]{0.5\linewidth}
        \adjincludegraphics[height=2.9cm,trim={{0.035\width} 0 {0.01\width} 0},clip]{paper_figs/stylized_imagenet/small_datasets_Finetuning_resnet50.pdf}
        \caption{\textbf{Full-network} ResNet-50}
    \end{subfigure}
    \caption{We compare standard, stylized and robust ImageNet models on standard transfer tasks.}
\end{figure}

\clearpage
\subsection{Unified scale: additional results to Figure~\ref{fig:downscaling-experiment}}

\begin{figure}[!htbp]
    \begin{subfigure}[]{\linewidth}
        \centering
        \includegraphics[width=0.7\linewidth]{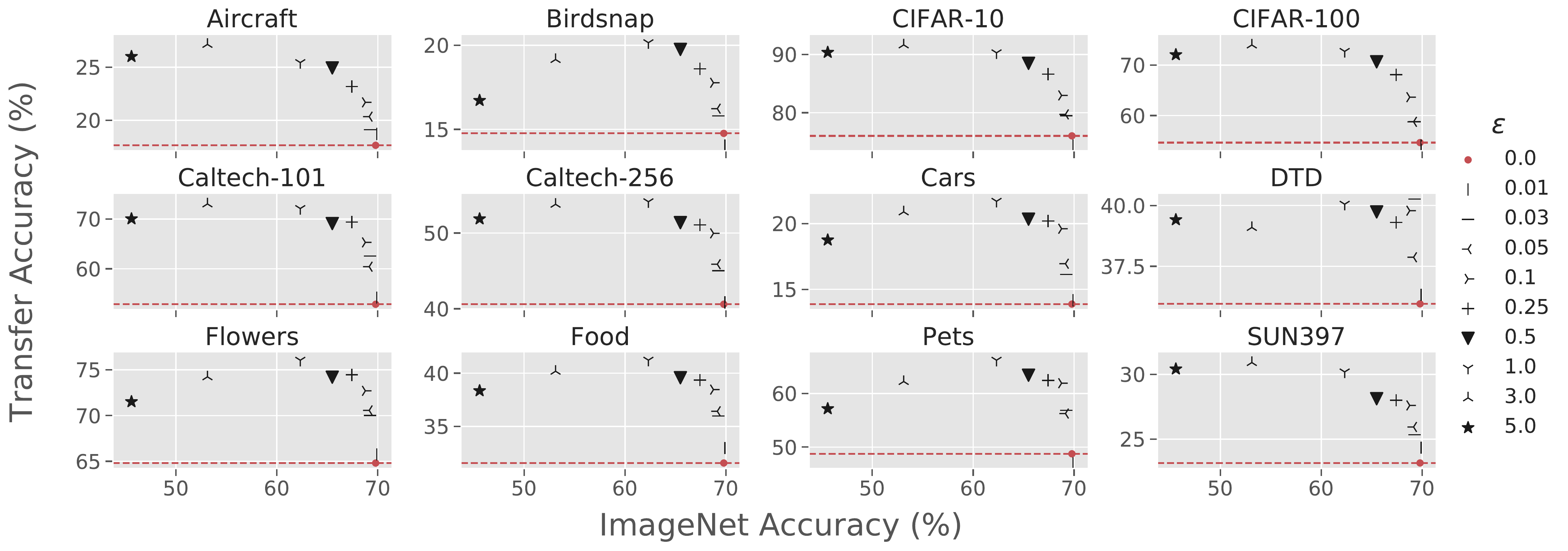}
        \caption{ResNet-18}
     \end{subfigure}
     \begin{subfigure}[]{\linewidth}
         \centering
         \includegraphics[width=0.7\linewidth]{paper_figs/unified_scale_4x3/resnet50_LogisticRegression.pdf}
        \caption{ResNet-50 (same as Figure~\ref{fig:downscaling-experiment})}
     \end{subfigure}
     \caption{\textbf{Fixed-feature} transfer accuracies of various datasets that
       are down-scaled to $32\times 32$ before being up-scaled again to ImageNet
       scale and used for transfer learning. The accuracy curves are closely
       aligned, unlike those of
       Figure~\ref{fig:main-small-datasets-transfer-results-logistic-regression},
       which illustrates the same experiment without downscaling.}
       \label{fig:downscaling-experiment_appendix_fixed_feature}
     \begin{subfigure}[]{\linewidth}
        \centering
       \includegraphics[width=0.7\linewidth]{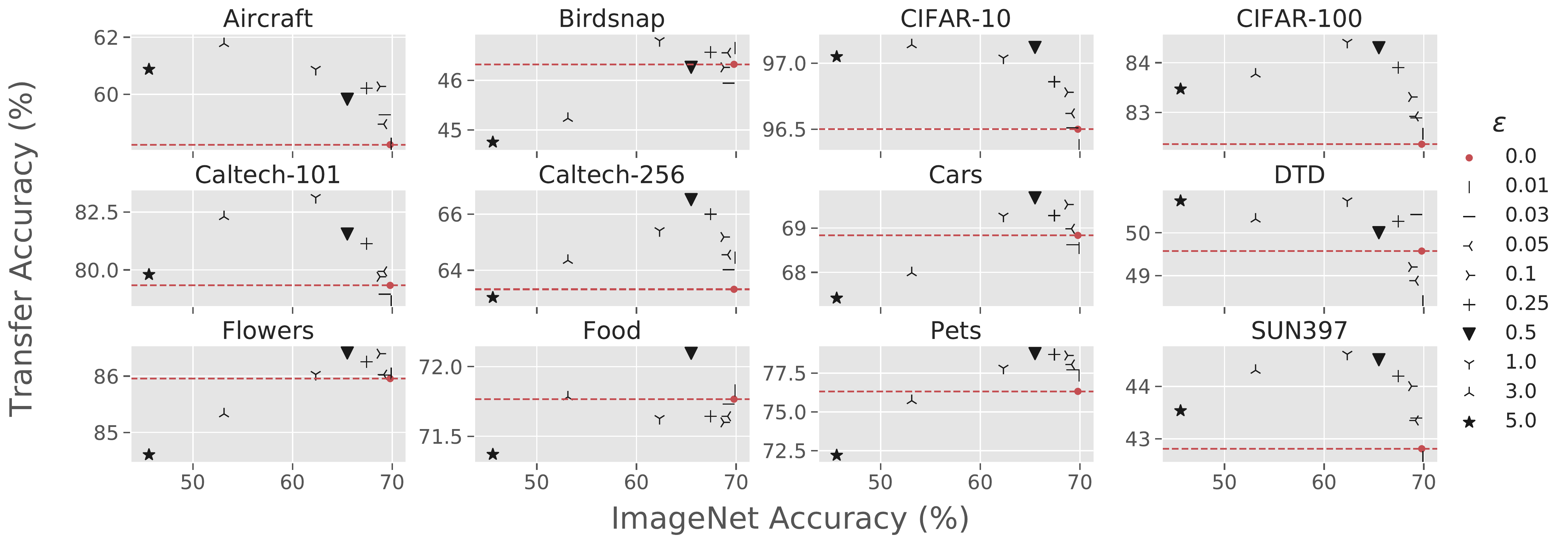}
       \caption{ResNet-18}
    \end{subfigure}
    \begin{subfigure}[]{\linewidth}
        \centering
        \includegraphics[width=0.7\linewidth]{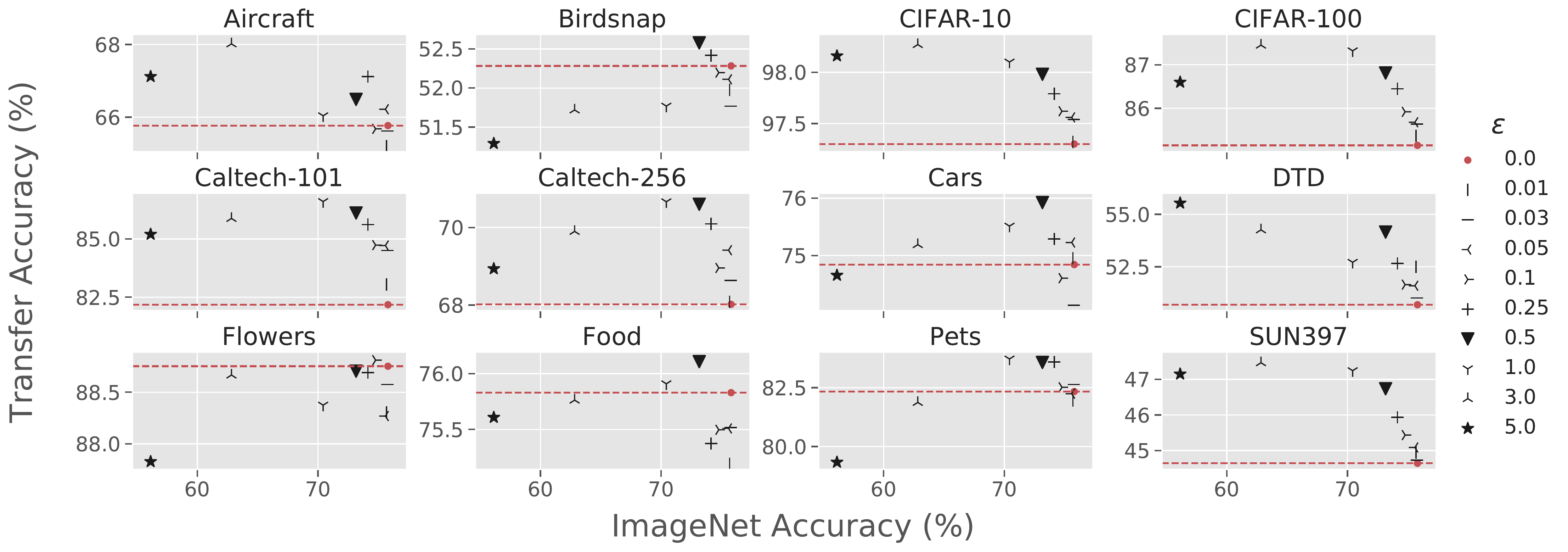}
       \caption{ResNet-50}
    \end{subfigure}
    \caption{\textbf{Full-network} transfer accuracies of various datasets that
      are down-scaled to $32\times 32$ before being up-scaled again to ImageNet
      scale and used for transfer learning.}
      \label{fig:downscaling-experiment_appendix_full_network}
\end{figure}

\clearpage
\subsection{Effect of width: additional results to Figure~\ref{fig:width-effect}}

\begin{figure}[!htbp]
    \centering
    \begin{subfigure}[]{.8\linewidth}
        \centering
        \adjincludegraphics[width=\linewidth]{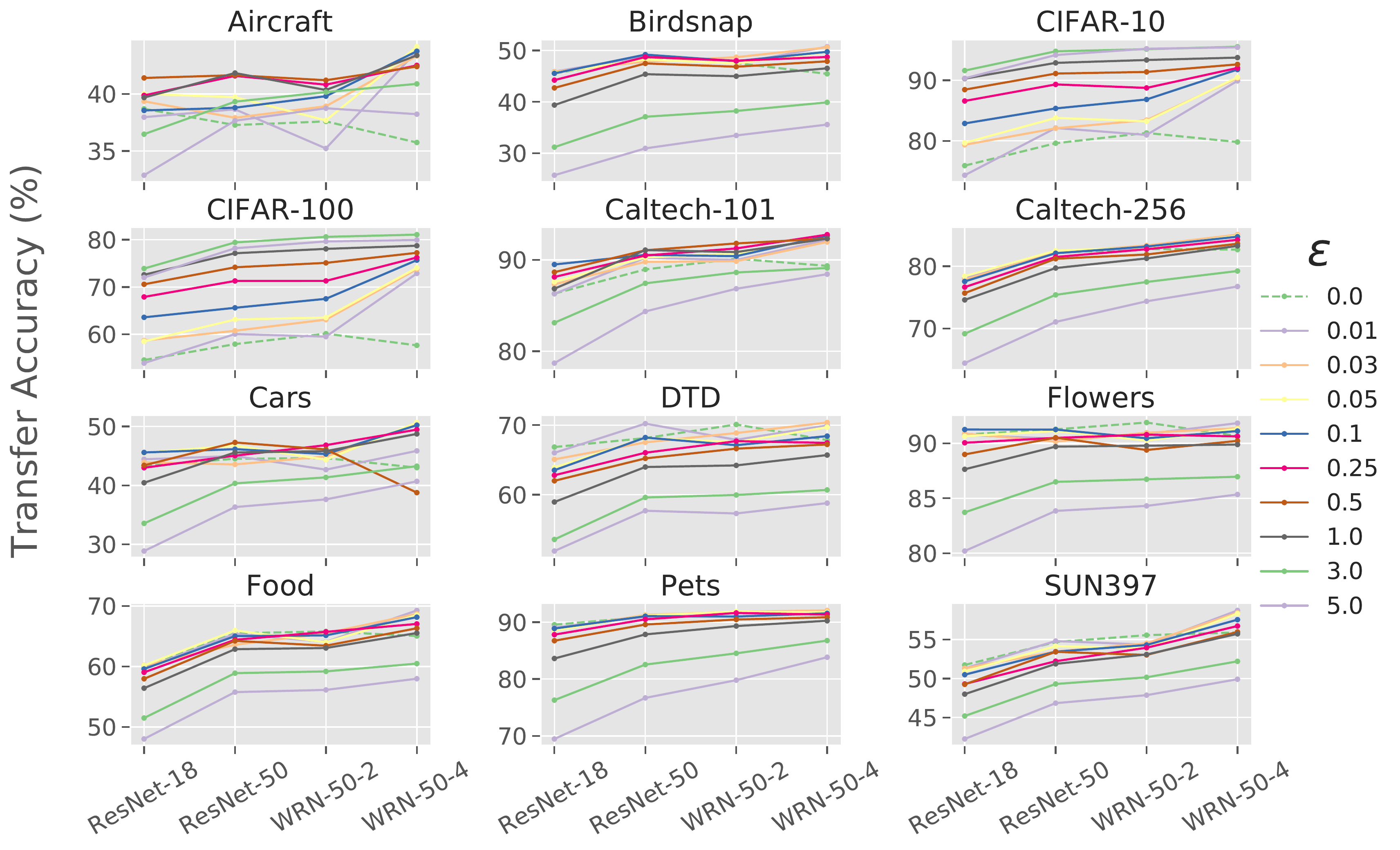}
        \caption{\textbf{Fixed-feature} transfer}
    \end{subfigure}
    \begin{subfigure}[]{.8\linewidth}
        \centering
        \adjincludegraphics[width=\linewidth]{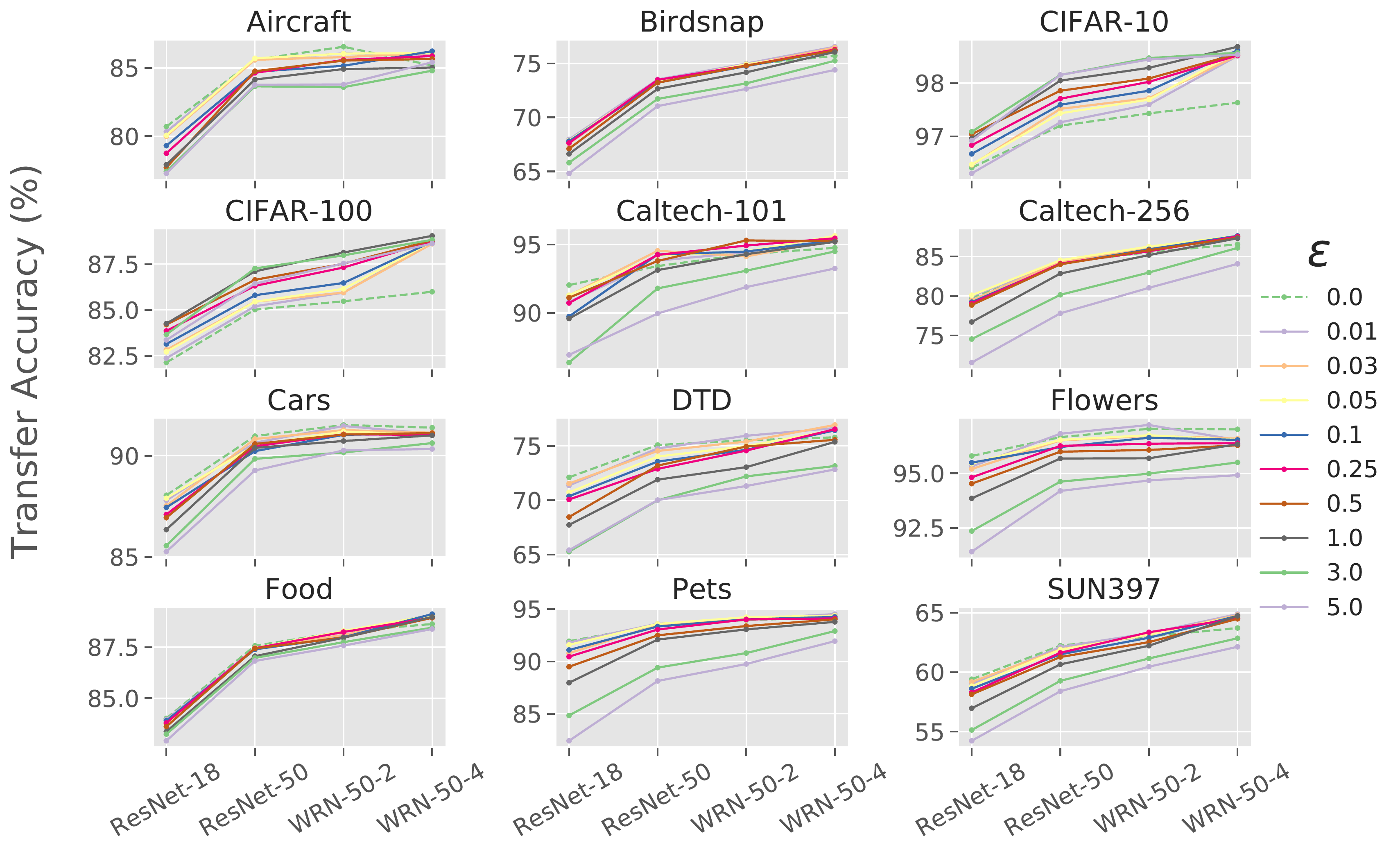}
        \caption{\textbf{Full-network} transfer}
    \end{subfigure}
    \caption{Varying width and model robustness while transfer learning from
      ImageNet to various datasets. Generally, as width increases, transfer
      learning accuracies of standard models generally plateau or level off
      while those of robust models steadily increase.}
    \label{fig:width-effect-appendix}
\end{figure}

%% file: sections/app_numerical_results.tex
\vskip -0.2cm
\subsection{Fixed-feature transfer to classification tasks (Fig. \ref{fig:main-small-datasets-transfer-results-logistic-regression})}
\vskip -0.5cm
\input{latex_tables/small_datasets_LogisticRegression}

\clearpage
\subsection{Full-network transfer to classification tasks (Fig. \ref{fig:main-small-datasets-transfer-results-finetuning})}
\input{latex_tables/small_datasets_Finetuning}

\clearpage

\subsection{Unifying dataset scale}
\subsubsection{Fixed-feature (cf. Fig. \ref{fig:downscaling-experiment} \& \ref{fig:downscaling-experiment_appendix_fixed_feature})}
\input{latex_tables/downsampling_LogisticRegression}

\clearpage
\subsubsection{Full-network (cf. Fig.~\ref{fig:downscaling-experiment_appendix_full_network})}
\input{latex_tables/downsampling_Finetuning}

%% file: latex_tables/small_datasets_LogisticRegression.tex
\begin{table}[!htbp]
\centering
\caption{\textbf{Fixed-feature} transfer for various standard and robust ImageNet models and datasets.}
\label{app:small-datasets-LogisticRegression}
\begin{small}
\resizebox*{!}{0.9\textheight}{
\begin{tabular}{llcccccccccc}
\toprule
       & {} & \multicolumn{10}{c}{\textbf{Transfer Accuracy (\%)}} \\\midrule
       && \multicolumn{10}{c}{Robustness parameter $\varepsilon$} \\
       \cmidrule(lr){3-12}
       &  &              0.00 &  0.01 &  0.03 &  0.05 &  0.10 &  0.25 &  0.50 &  1.00 &  3.00 &  5.00 \\
\textbf{Dataset} & \textbf{Model} &                   &       &       &       &       &       &       &       &       &       \\
\midrule
\multirow{4}{*}{\textbf{Aircraft}} & \textbf{ResNet-18} &             38.69 & 37.96 & 39.35 & 40.00 & 38.55 & 39.87 & \textbf{41.40} & 39.68 & 36.47 & 32.87 \\
       & \textbf{ResNet-50} &             37.27 & 38.65 & 37.91 & 39.71 & 38.79 & 41.58 & 41.64 & \textbf{41.83} & 39.32 & 37.65 \\
       & \textbf{WRN-50-2} &             37.59 & 35.22 & 38.92 & 37.68 & 39.80 & 40.81 & \textbf{41.20} & 40.34 & 40.16 & 38.74 \\
       & \textbf{WRN-50-4} &             35.74 & 43.76 & 43.34 & \textbf{44.14} & 43.75 & 42.51 & 42.40 & 43.38 & 40.88 & 38.23 \\
\midrule
\multirow{4}{*}{\textbf{Birdsnap}} & \textbf{ResNet-18} &             45.54 & \textbf{45.88} & 45.86 & 45.66 & 45.55 & 44.23 & 42.72 & 39.38 & 31.19 & 25.73 \\
       & \textbf{ResNet-50} &             48.35 & 48.86 & 47.84 & 48.24 & \textbf{49.19} & 48.73 & 47.48 & 45.38 & 37.10 & 30.95 \\
       & \textbf{WRN-50-2} &             47.54 & 47.47 & \textbf{48.68} & 47.48 & 47.93 & 48.01 & 46.84 & 44.99 & 38.23 & 33.47 \\
       & \textbf{WRN-50-4} &             45.45 & \textbf{50.72} & 50.60 & 49.66 & 49.73 & 48.73 & 47.88 & 46.53 & 39.91 & 35.58 \\
\midrule
\multirow{4}{*}{\textbf{CIFAR-10}} & \textbf{ResNet-18} &             75.91 & 74.33 & 79.35 & 79.67 & 82.87 & 86.58 & 88.45 & 90.27 & \textbf{91.59} & 90.31 \\
       & \textbf{ResNet-50} &             79.61 & 82.12 & 82.07 & 83.78 & 85.35 & 89.31 & 91.10 & 92.86 & \textbf{94.77} & 94.16 \\
       & \textbf{WRN-50-2} &             81.31 & 80.98 & 83.43 & 83.23 & 86.83 & 88.73 & 91.37 & 93.34 & 95.12 & \textbf{95.19} \\
       & \textbf{WRN-50-4} &             79.81 & 89.90 & 90.35 & 90.48 & 91.76 & 92.03 & 92.62 & 93.73 & \textbf{95.53} & 95.43 \\
\midrule
\multirow{4}{*}{\textbf{CIFAR-100}} & \textbf{ResNet-18} &             54.58 & 53.92 & 58.70 & 58.51 & 63.60 & 67.91 & 70.58 & 72.60 & \textbf{73.91} & 72.01 \\
       & \textbf{ResNet-50} &             57.94 & 60.06 & 60.76 & 63.13 & 65.61 & 71.29 & 74.18 & 77.14 & \textbf{79.43} & 78.20 \\
       & \textbf{WRN-50-2} &             60.14 & 59.52 & 63.12 & 63.55 & 67.51 & 71.30 & 75.11 & 78.07 & \textbf{80.61} & 79.64 \\
       & \textbf{WRN-50-4} &             57.68 & 72.88 & 73.79 & 74.06 & 75.68 & 76.25 & 77.23 & 78.73 & \textbf{81.08} & 79.94 \\
\midrule
\multirow{4}{*}{\textbf{Caltech-101}} & \textbf{ResNet-18} &             86.30 & 86.28 & 87.32 & 87.59 & \textbf{89.49} & 88.12 & 88.65 & 86.84 & 83.11 & 78.69 \\
       & \textbf{ResNet-50} &             88.95 & 90.22 & 89.79 & 90.26 & 90.54 & 90.48 & 91.04 & \textbf{91.07} & 87.43 & 84.35 \\
       & \textbf{WRN-50-2} &             90.12 & 89.97 & 89.85 & 90.67 & 90.40 & 91.25 & \textbf{91.80} & 90.84 & 88.62 & 86.83 \\
       & \textbf{WRN-50-4} &             89.34 & 92.20 & 91.96 & 92.44 & 92.63 & \textbf{92.76} & 92.32 & 92.32 & 89.10 & 88.43 \\
\midrule
\multirow{4}{*}{\textbf{Caltech-256}} & \textbf{ResNet-18} &             77.58 & 78.09 & 77.87 & \textbf{78.40} & 77.57 & 76.66 & 75.69 & 74.61 & 69.19 & 64.46 \\
       & \textbf{ResNet-50} &             82.21 & 82.31 & 82.23 & \textbf{82.51} & 82.10 & 81.50 & 81.21 & 79.72 & 75.42 & 71.07 \\
       & \textbf{WRN-50-2} &             82.78 & 82.94 & \textbf{83.34} & 83.04 & 83.17 & 82.74 & 81.89 & 81.26 & 77.48 & 74.38 \\
       & \textbf{WRN-50-4} &             82.68 & 85.07 & \textbf{85.08} & 84.88 & 84.75 & 84.24 & 83.62 & 83.27 & 79.24 & 76.75 \\
\midrule
\multirow{4}{*}{\textbf{Cars}} & \textbf{ResNet-18} &             43.34 & 44.43 & 43.92 & 45.53 & \textbf{45.59} & 43.00 & 43.40 & 40.45 & 33.55 & 28.86 \\
       & \textbf{ResNet-50} &             44.52 & 44.98 & 43.56 & 46.74 & 46.15 & 45.04 & \textbf{47.28} & 45.58 & 40.34 & 36.32 \\
       & \textbf{WRN-50-2} &             44.63 & 42.67 & 44.92 & 44.36 & 45.32 & \textbf{46.83} & 46.10 & 45.81 & 41.35 & 37.62 \\
       & \textbf{WRN-50-4} &             43.01 & 45.86 & 50.39 & \textbf{50.67} & 50.22 & 49.46 & 38.77 & 48.73 & 43.26 & 40.68 \\
\midrule
\multirow{4}{*}{\textbf{DTD}} & \textbf{ResNet-18} &             \textbf{66.84} & 66.01 & 65.07 & 63.90 & 63.51 & 62.78 & 61.99 & 58.94 & 53.55 & 51.88 \\
       & \textbf{ResNet-50} &             68.14 & \textbf{70.21} & 67.52 & 68.16 & 68.21 & 66.03 & 65.21 & 63.97 & 59.59 & 57.68 \\
       & \textbf{WRN-50-2} &             \textbf{70.09} & 67.89 & 68.87 & 67.55 & 67.11 & 67.70 & 66.61 & 64.20 & 59.95 & 57.29 \\
       & \textbf{WRN-50-4} &             67.85 & 69.95 & \textbf{70.37} & 69.70 & 68.42 & 67.45 & 67.22 & 65.69 & 60.67 & 58.78 \\
\midrule
\multirow{4}{*}{\textbf{Flowers}} & \textbf{ResNet-18} &             90.80 & 90.76 & 90.88 & 90.65 & \textbf{91.26} & 90.05 & 88.99 & 87.64 & 83.72 & 80.20 \\
       & \textbf{ResNet-50} &             \textbf{91.28} & 90.43 & 90.16 & 91.12 & 91.26 & 90.50 & 90.52 & 89.70 & 86.49 & 83.85 \\
       & \textbf{WRN-50-2} &             \textbf{91.90} & 90.86 & 90.97 & 90.26 & 90.46 & 90.79 & 89.39 & 89.79 & 86.73 & 84.31 \\
       & \textbf{WRN-50-4} &             90.67 & \textbf{91.84} & 91.37 & 91.32 & 91.12 & 90.63 & 90.23 & 89.89 & 86.96 & 85.35 \\
\midrule
\multirow{4}{*}{\textbf{Food}} & \textbf{ResNet-18} &             59.96 & 59.67 & \textbf{60.20} & 60.17 & 59.59 & 59.04 & 57.97 & 56.42 & 51.49 & 48.03 \\
       & \textbf{ResNet-50} &             65.49 & 65.39 & 63.59 & \textbf{65.95} & 65.02 & 64.41 & 64.23 & 62.86 & 58.90 & 55.77 \\
       & \textbf{WRN-50-2} &             \textbf{65.80} & 64.06 & 65.50 & 64.00 & 65.14 & 65.73 & 63.44 & 63.05 & 59.19 & 56.13 \\
       & \textbf{WRN-50-4} &             65.04 & \textbf{69.26} & 68.69 & 68.50 & 68.15 & 67.03 & 66.32 & 65.53 & 60.48 & 57.98 \\
\midrule
\multirow{4}{*}{\textbf{Pets}} & \textbf{ResNet-18} &             \textbf{89.55} & 89.03 & 88.67 & 88.54 & 88.87 & 87.80 & 86.73 & 83.61 & 76.29 & 69.48 \\
       & \textbf{ResNet-50} &             90.92 & 90.93 & \textbf{91.27} & 91.16 & 91.05 & 90.48 & 89.57 & 87.84 & 82.54 & 76.69 \\
       & \textbf{WRN-50-2} &             91.81 & 91.69 & 91.83 & \textbf{91.85} & 90.98 & 91.61 & 90.46 & 89.31 & 84.51 & 79.80 \\
       & \textbf{WRN-50-4} &             91.83 & 91.82 & \textbf{92.05} & 91.70 & 91.54 & 91.32 & 90.85 & 90.23 & 86.75 & 83.83 \\
\midrule
\multirow{4}{*}{\textbf{SUN397}} & \textbf{ResNet-18} &             \textbf{51.74} & 51.31 & 51.32 & 50.92 & 50.50 & 49.30 & 49.25 & 47.99 & 45.19 & 42.24 \\
       & \textbf{ResNet-50} &             54.69 & \textbf{54.82} & 53.48 & 54.15 & 53.45 & 52.23 & 53.43 & 51.88 & 49.30 & 46.84 \\
       & \textbf{WRN-50-2} &             \textbf{55.57} & 54.35 & 54.53 & 53.90 & 54.31 & 53.96 & 53.03 & 53.09 & 50.16 & 47.86 \\
       & \textbf{WRN-50-4} &             55.92 & \textbf{58.75} & 58.45 & 58.34 & 57.56 & 56.75 & 55.99 & 55.74 & 52.21 & 49.91 \\
\bottomrule
\end{tabular}
}
\end{small}
\end{table}

%% file: latex_tables/small_datasets_Finetuning.tex
\begin{table}[!htbp]
\centering
\caption{\textbf{Full-network} transfer for various standard and robust ImageNet models and datasets.}
\label{app:small-datasets-Finetuning}
\begin{small}
\resizebox*{!}{0.9\textheight}{
\begin{tabular}{llcccccccccc}
\toprule
       & {} & \multicolumn{10}{c}{\textbf{Transfer Accuracy (\%)}} \\\midrule
       && \multicolumn{10}{c}{Robustness parameter $\varepsilon$} \\
       \cmidrule(lr){3-12}
       &  &              0.00 &  0.01 &  0.03 &  0.05 &  0.10 &  0.25 &  0.50 &  1.00 &  3.00 &  5.00 \\
\textbf{Dataset} & \textbf{Model} &                   &       &       &       &       &       &       &       &       &       \\
\midrule
\multirow{4}{*}{\textbf{Aircraft}} & \textbf{ResNet-18} &             \textbf{80.70} & 80.32 & 79.99 & 80.06 & 79.30 & 78.74 & 77.69 & 77.90 & 77.41 & 77.26 \\
       & \textbf{ResNet-50} &             85.62 & 85.62 & 85.61 & \textbf{85.72} & 84.73 & 84.65 & 84.77 & 84.16 & 83.66 & 83.77 \\
       & \textbf{WRN-50-2} &             \textbf{86.57} & 86.08 & 85.81 & 86.06 & 85.17 & 85.60 & 85.55 & 84.93 & 83.60 & 83.80 \\
       & \textbf{WRN-50-4} &             85.19 & 85.98 & 86.10 & 86.11 & \textbf{86.24} & 85.88 & 85.67 & 85.04 & 84.81 & 85.43 \\
\midrule
\multirow{4}{*}{\textbf{Birdsnap}} & \textbf{ResNet-18} &             67.71 & \textbf{67.96} & 67.58 & 67.86 & 67.80 & 67.63 & 67.10 & 66.62 & 65.80 & 64.81 \\
       & \textbf{ResNet-50} &             73.38 & \textbf{73.52} & 73.39 & 73.33 & 73.22 & 73.48 & 73.21 & 72.65 & 71.71 & 71.05 \\
       & \textbf{WRN-50-2} &             74.87 & \textbf{74.98} & 74.85 & 74.93 & 74.75 & 74.80 & 74.79 & 74.18 & 73.15 & 72.64 \\
       & \textbf{WRN-50-4} &             75.71 & \textbf{76.55} & 76.47 & 76.14 & 76.18 & 76.29 & 76.20 & 76.06 & 75.25 & 74.40 \\
\midrule
\multirow{4}{*}{\textbf{CIFAR-10}} & \textbf{ResNet-18} &             96.41 & 96.30 & 96.46 & 96.47 & 96.67 & 96.83 & 97.04 & 96.96 & \textbf{97.09} & 96.92 \\
       & \textbf{ResNet-50} &             97.20 & 97.26 & 97.52 & 97.43 & 97.59 & 97.71 & 97.86 & 98.05 & \textbf{98.15} & \textbf{98.15} \\
       & \textbf{WRN-50-2} &             97.43 & 97.60 & 97.72 & 97.69 & 97.86 & 98.02 & 98.09 & 98.29 & \textbf{98.47} & 98.44 \\
       & \textbf{WRN-50-4} &             97.63 & 98.51 & 98.52 & 98.59 & 98.62 & 98.52 & 98.55 & \textbf{98.68} & 98.57 & 98.53 \\
\midrule
\multirow{4}{*}{\textbf{CIFAR-100}} & \textbf{ResNet-18} &             82.13 & 82.36 & 82.82 & 82.71 & 83.14 & 83.85 & 84.19 & \textbf{84.25} & 83.65 & 83.36 \\
       & \textbf{ResNet-50} &             85.02 & 85.20 & 85.45 & 85.44 & 85.80 & 86.31 & 86.64 & 87.10 & \textbf{87.26} & 86.43 \\
       & \textbf{WRN-50-2} &             85.47 & 85.94 & 85.95 & 86.15 & 86.47 & 87.31 & 87.52 & \textbf{88.13} & 87.98 & 87.54 \\
       & \textbf{WRN-50-4} &             85.99 & 88.70 & 88.61 & 88.72 & 88.72 & 88.75 & 88.80 & \textbf{89.04} & 88.83 & 88.62 \\
\midrule
\multirow{4}{*}{\textbf{Caltech-101}} & \textbf{ResNet-18} &             \textbf{92.04} & 90.81 & 91.28 & 91.29 & 89.75 & 90.73 & 91.12 & 89.60 & 86.39 & 86.95 \\
       & \textbf{ResNet-50} &             93.42 & 93.82 & \textbf{94.53} & 94.18 & 94.27 & 94.24 & 93.79 & 93.13 & 91.79 & 89.97 \\
       & \textbf{WRN-50-2} &             94.29 & 94.43 & 94.13 & 94.49 & 94.48 & 94.92 & \textbf{95.29} & 94.28 & 93.08 & 91.89 \\
       & \textbf{WRN-50-4} &             94.76 & 95.60 & 95.32 & \textbf{95.62} & 95.30 & 95.45 & 95.23 & 95.19 & 94.49 & 93.25 \\
\midrule
\multirow{4}{*}{\textbf{Caltech-256}} & \textbf{ResNet-18} &             79.80 & 80.00 & 79.45 & \textbf{80.10} & 79.23 & 79.07 & 78.86 & 76.71 & 74.55 & 71.57 \\
       & \textbf{ResNet-50} &             84.19 & 84.30 & 84.37 & \textbf{84.54} & 84.04 & 84.12 & 84.02 & 82.85 & 80.15 & 77.81 \\
       & \textbf{WRN-50-2} &             85.56 & 85.65 & 86.04 & \textbf{86.26} & 85.91 & 85.67 & 85.80 & 85.19 & 82.97 & 81.04 \\
       & \textbf{WRN-50-4} &             86.56 & 87.53 & 87.54 & \textbf{87.62} & \textbf{87.62} & 87.54 & 87.38 & 87.31 & 86.09 & 84.08 \\
\midrule
\multirow{4}{*}{\textbf{Cars}} & \textbf{ResNet-18} &             \textbf{88.05} & 87.80 & 87.53 & 87.90 & 87.45 & 87.10 & 86.94 & 86.35 & 85.56 & 85.26 \\
       & \textbf{ResNet-50} &             \textbf{90.97} & 90.65 & 90.83 & 90.52 & 90.23 & 90.47 & 90.59 & 90.39 & 89.85 & 89.28 \\
       & \textbf{WRN-50-2} &             \textbf{91.52} & 91.47 & 91.27 & 91.20 & 91.04 & 91.06 & 91.05 & 90.73 & 90.16 & 90.27 \\
       & \textbf{WRN-50-4} &             \textbf{91.39} & 91.09 & 91.14 & 91.05 & 91.10 & 91.03 & 91.12 & 91.01 & 90.63 & 90.34 \\
\midrule
\multirow{4}{*}{\textbf{DTD}} & \textbf{ResNet-18} &             \textbf{72.11} & 71.37 & 71.54 & 70.73 & 70.37 & 70.07 & 68.46 & 67.73 & 65.27 & 65.41 \\
       & \textbf{ResNet-50} &             \textbf{75.09} & 74.77 & 74.54 & 74.02 & 73.56 & 72.89 & 73.19 & 71.90 & 70.00 & 70.02 \\
       & \textbf{WRN-50-2} &             75.51 & \textbf{75.94} & 75.41 & 74.98 & 74.65 & 74.57 & 74.95 & 73.05 & 72.20 & 71.31 \\
       & \textbf{WRN-50-4} &             75.80 & 76.65 & \textbf{76.93} & 76.47 & 76.44 & 76.54 & 75.57 & 75.37 & 73.16 & 72.84 \\
\midrule
\multirow{4}{*}{\textbf{Flowers}} & \textbf{ResNet-18} &             \textbf{95.79} & 95.31 & 95.20 & 95.44 & 95.49 & 94.82 & 94.53 & 93.86 & 92.36 & 91.42 \\
       & \textbf{ResNet-50} &             96.65 & \textbf{96.81} & 96.50 & 96.53 & 96.20 & 96.25 & 95.99 & 95.68 & 94.62 & 94.20 \\
       & \textbf{WRN-50-2} &             97.04 & \textbf{97.21} & 96.71 & 96.74 & 96.63 & 96.35 & 96.07 & 95.69 & 94.98 & 94.67 \\
       & \textbf{WRN-50-4} &             \textbf{97.01} & 96.52 & 96.59 & 96.53 & 96.53 & 96.38 & 96.28 & 96.33 & 95.50 & 94.92 \\
\midrule
\multirow{4}{*}{\textbf{Food}} & \textbf{ResNet-18} &             \textbf{84.01} & 83.95 & 83.74 & 83.69 & 83.89 & 83.78 & 83.60 & 83.36 & 83.23 & 82.91 \\
       & \textbf{ResNet-50} &             \textbf{87.57} & 87.42 & 87.45 & 87.46 & 87.40 & 87.45 & 87.44 & 87.06 & 86.97 & 86.82 \\
       & \textbf{WRN-50-2} &             88.27 & 88.26 & 88.10 & \textbf{88.30} & 87.99 & 88.25 & 87.97 & 87.96 & 87.75 & 87.58 \\
       & \textbf{WRN-50-4} &             88.64 & 89.09 & 89.00 & 89.08 & \textbf{89.12} & 88.95 & 88.94 & 88.98 & 88.46 & 88.39 \\
\midrule
\multirow{4}{*}{\textbf{Pets}} & \textbf{ResNet-18} &             \textbf{91.94} & 91.81 & 90.79 & 91.59 & 91.09 & 90.46 & 89.49 & 87.96 & 84.83 & 82.41 \\
       & \textbf{ResNet-50} &             93.49 & \textbf{93.61} & 93.50 & 93.59 & 93.34 & 93.06 & 92.50 & 92.09 & 89.41 & 88.13 \\
       & \textbf{WRN-50-2} &             93.96 & 94.05 & 93.98 & \textbf{94.23} & 94.02 & 94.02 & 93.39 & 93.07 & 90.80 & 89.76 \\
       & \textbf{WRN-50-4} &             94.20 & \textbf{94.53} & 94.40 & 94.38 & 94.27 & 94.11 & 94.02 & 93.79 & 92.91 & 91.94 \\
\midrule
\multirow{4}{*}{\textbf{SUN397}} & \textbf{ResNet-18} &             \textbf{59.41} & 58.98 & 59.19 & 58.83 & 58.61 & 58.29 & 58.14 & 56.97 & 55.14 & 54.23 \\
       & \textbf{ResNet-50} &             \textbf{62.24} & 62.12 & 61.93 & 61.89 & 61.50 & 61.64 & 61.28 & 60.66 & 59.27 & 58.40 \\
       & \textbf{WRN-50-2} &             63.02 & 63.28 & 63.16 & 63.18 & 62.90 & \textbf{63.36} & 62.53 & 62.23 & 61.16 & 60.47 \\
       & \textbf{WRN-50-4} &             63.72 & \textbf{64.89} & 64.81 & 64.71 & 64.74 & 64.53 & 64.49 & 64.74 & 62.86 & 62.14 \\
\bottomrule
\end{tabular}
}
\end{small}
\end{table}

%% file: latex_tables/downsampling_LogisticRegression.tex
\begin{table}[!htbp]
\centering
\caption{\textbf{Fixed-feature} transfer on 32x32 downsampled datasets.}
\label{app:downsampling-LogisticRegression}
\begin{small}
\begin{tabular}{llcccccccccc}
\toprule
       & {} & \multicolumn{10}{c}{\textbf{Transfer Accuracy (\%)}} \\\midrule
       && \multicolumn{10}{c}{Robustness parameter $\varepsilon$} \\
       \cmidrule(lr){3-12}
       &&              0.00 &  0.01 &  0.03 &  0.05 &  0.10 &  0.25 &  0.50 &  1.00 &  3.00 &  5.00 \\
\textbf{Dataset} & \textbf{Model} &                   &       &       &       &       &       &       &       &       &       \\
\midrule
\multirow{2}{*}{\textbf{Aircraft}} & \textbf{ResNet-18} &             17.64 & 18.72 & 19.11 & 20.34 & 21.69 & 23.19 & 24.93 & 25.44 & \textbf{27.15} & 26.01 \\
       & \textbf{ResNet-50} &             15.87 & 17.04 & 17.82 & 18.48 & 20.19 & 22.44 & 24.12 & 25.89 & \textbf{28.59} & 28.35 \\
\midrule
\multirow{2}{*}{\textbf{Birdsnap}} & \textbf{ResNet-18} &             14.76 & 14.04 & 15.80 & 16.23 & 17.77 & 18.60 & 19.75 & \textbf{20.16} & 19.15 & 16.72 \\
       & \textbf{ResNet-50} &             13.85 & 14.12 & 14.67 & 15.42 & 16.94 & 19.67 & 21.74 & \textbf{23.08} & 22.98 & 20.70 \\
\midrule
\multirow{2}{*}{\textbf{CIFAR-10}} & \textbf{ResNet-18} &             76.02 & 74.36 & 79.48 & 79.71 & 82.97 & 86.62 & 88.47 & 90.29 & \textbf{91.64} & 90.36 \\
       & \textbf{ResNet-50} &             79.63 & 82.18 & 82.15 & 83.88 & 85.41 & 89.35 & 91.13 & 92.89 & \textbf{94.81} & 94.23 \\
\midrule
\multirow{2}{*}{\textbf{CIFAR-100}} & \textbf{ResNet-18} &             54.61 & 54.03 & 58.77 & 58.74 & 63.64 & 68.10 & 70.66 & 72.74 & \textbf{74.01} & 72.08 \\
       & \textbf{ResNet-50} &             58.01 & 60.17 & 60.87 & 63.24 & 65.73 & 71.32 & 74.19 & 77.17 & \textbf{79.50} & 78.27 \\
\midrule
\multirow{2}{*}{\textbf{Caltech-101}} & \textbf{ResNet-18} &             52.88 & 54.20 & 62.56 & 60.43 & 65.31 & 69.39 & 69.08 & 72.11 & \textbf{73.02} & 70.04 \\
       & \textbf{ResNet-50} &             56.55 & 59.32 & 60.45 & 61.08 & 63.76 & 69.80 & 73.11 & 76.89 & \textbf{78.86} & 77.43 \\
\midrule
\multirow{2}{*}{\textbf{Caltech-256}} & \textbf{ResNet-18} &             40.60 & 40.83 & 45.02 & 45.88 & 49.96 & 51.08 & 51.36 & \textbf{54.13} & 53.79 & 51.87 \\
       & \textbf{ResNet-50} &             42.73 & 45.11 & 45.65 & 47.52 & 49.61 & 53.63 & 56.12 & 58.93 & \textbf{59.79} & 58.67 \\
\midrule
\multirow{2}{*}{\textbf{Cars}} & \textbf{ResNet-18} &             13.88 & 14.18 & 16.14 & 16.95 & 19.61 & 20.20 & 20.33 & \textbf{21.70} & 20.89 & 18.75 \\
       & \textbf{ResNet-50} &             13.16 & 13.89 & 13.68 & 16.84 & 17.07 & 19.40 & 21.88 & 23.19 & \textbf{24.19} & 23.37 \\
\midrule
\multirow{2}{*}{\textbf{DTD}} & \textbf{ResNet-18} &             35.96 & 36.33 & \textbf{40.27} & 37.87 & 39.79 & 39.31 & 39.73 & 40.05 & 39.10 & 39.41 \\
       & \textbf{ResNet-50} &             41.28 & 40.37 & 41.06 & 42.13 & 41.22 & 43.56 & 44.10 & 43.78 & 43.83 & \textbf{44.26} \\
\midrule
\multirow{2}{*}{\textbf{Flowers}} & \textbf{ResNet-18} &             64.81 & 65.75 & 70.01 & 70.57 & 72.71 & 74.46 & 74.19 & \textbf{76.06} & 74.23 & 71.52 \\
       & \textbf{ResNet-50} &             66.65 & 68.49 & 68.24 & 71.03 & 73.12 & 75.83 & 76.52 & 77.23 & \textbf{78.31} & 75.71 \\
\midrule
\multirow{2}{*}{\textbf{Food}} & \textbf{ResNet-18} &             31.58 & 32.98 & 35.98 & 36.42 & 38.46 & 39.35 & 39.56 & \textbf{41.22} & 40.17 & 38.35 \\
       & \textbf{ResNet-50} &             36.46 & 36.82 & 36.37 & 39.85 & 40.91 & 43.08 & 44.88 & 46.16 & \textbf{46.45} & 44.44 \\
\midrule
\multirow{2}{*}{\textbf{Pets}} & \textbf{ResNet-18} &             48.74 & 46.98 & 56.87 & 56.25 & 61.92 & 62.45 & 63.39 & \textbf{66.20} & 62.23 & 57.15 \\
       & \textbf{ResNet-50} &             53.98 & 54.10 & 58.55 & 53.57 & 59.58 & 67.35 & 69.31 & \textbf{70.16} & 69.43 & 64.37 \\
\midrule
\multirow{2}{*}{\textbf{SUN397}} & \textbf{ResNet-18} &             23.16 & 24.35 & 25.34 & 25.94 & 27.60 & 28.00 & 28.12 & 30.19 & \textbf{30.91} & 30.41 \\
       & \textbf{ResNet-50} &             23.62 & 25.60 & 24.64 & 27.30 & 27.56 & 29.24 & 31.36 & 32.37 & \textbf{33.90} & 33.58 \\
\bottomrule
\end{tabular}
\end{small}
\end{table}

%% file: latex_tables/downsampling_Finetuning.tex
\begin{table}[!htbp]
\centering
\caption{\textbf{Full-network} transfer on 32x32 downsampled datasets.}
\label{app:downsampling-Finetuning}
\begin{small}
\begin{tabular}{llcccccccccc}
\toprule
       & {} & \multicolumn{10}{c}{\textbf{Transfer Accuracy (\%)}} \\\midrule
       && \multicolumn{10}{c}{Robustness parameter $\varepsilon$} \\
       \cmidrule(lr){3-12}
       &&              0.00 &  0.01 &  0.03 &  0.05 &  0.10 &  0.25 &  0.50 &  1.00 &  3.00 &  5.00 \\
\textbf{Dataset} & \textbf{Model} &                   &       &       &       &       &       &       &       &       &       \\
\midrule
\multirow{2}{*}{\textbf{Aircraft}} & \textbf{ResNet-18} &             58.24 & 58.27 & 59.29 & 58.96 & 60.28 & 60.22 & 59.83 & 60.88 & \textbf{61.78} & 60.88 \\
       & \textbf{ResNet-50} &             65.77 & 65.20 & 65.62 & 66.22 & 65.68 & 67.12 & 66.49 & 66.04 & \textbf{68.02} & 67.12 \\
\midrule
\multirow{2}{*}{\textbf{Birdsnap}} & \textbf{ResNet-18} &             46.32 & 46.65 & 45.94 & 46.55 & 46.26 & 46.57 & 46.26 & \textbf{46.80} & 45.23 & 44.76 \\
       & \textbf{ResNet-50} &             52.28 & 51.98 & 51.77 & 52.11 & 52.20 & 52.42 & \textbf{52.58} & 51.77 & 51.72 & 51.29 \\
\midrule
\multirow{2}{*}{\textbf{CIFAR-10}} & \textbf{ResNet-18} &             96.50 & 96.38 & 96.51 & 96.62 & 96.78 & 96.86 & 97.12 & 97.04 & \textbf{97.14} & 97.05 \\
       & \textbf{ResNet-50} &             97.30 & 97.32 & 97.54 & 97.56 & 97.62 & 97.79 & 97.98 & 98.10 & \textbf{98.27} & 98.16 \\
\midrule
\multirow{2}{*}{\textbf{CIFAR-100}} & \textbf{ResNet-18} &             82.36 & 82.57 & 82.89 & 82.92 & 83.31 & 83.90 & 84.30 & \textbf{84.41} & 83.77 & 83.47 \\
       & \textbf{ResNet-50} &             85.15 & 85.37 & 85.64 & 85.68 & 85.92 & 86.45 & 86.81 & 87.32 & \textbf{87.45} & 86.60 \\
\midrule
\multirow{2}{*}{\textbf{Caltech-101}} & \textbf{ResNet-18} &             79.33 & 78.64 & 78.95 & 79.94 & 79.70 & 81.13 & 81.55 & \textbf{83.13} & 82.30 & 79.80 \\
       & \textbf{ResNet-50} &             82.18 & 83.05 & 84.50 & 84.72 & 84.74 & 85.62 & 86.12 & \textbf{86.61} & 85.88 & 85.20 \\
\midrule
\multirow{2}{*}{\textbf{Caltech-256}} & \textbf{ResNet-18} &             63.32 & 64.45 & 64.02 & 64.55 & 65.18 & 66.00 & \textbf{66.52} & 65.41 & 64.35 & 63.03 \\
       & \textbf{ResNet-50} &             68.02 & 68.09 & 68.63 & 69.42 & 68.96 & 70.10 & 70.60 & \textbf{70.66} & 69.90 & 68.94 \\
\midrule
\multirow{2}{*}{\textbf{Cars}} & \textbf{ResNet-18} &             68.83 & 68.55 & 68.62 & 68.98 & 69.53 & 69.28 & \textbf{69.68} & 69.27 & 67.99 & 67.42 \\
       & \textbf{ResNet-50} &             74.84 & 74.95 & 74.13 & 75.23 & 74.61 & 75.29 & \textbf{75.92} & 75.51 & 75.19 & 74.65 \\
\midrule
\multirow{2}{*}{\textbf{DTD}} & \textbf{ResNet-18} &             49.57 & 48.40 & 50.43 & 48.88 & 49.20 & 50.27 & 50.00 & \textbf{50.74} & 50.32 & \textbf{50.74} \\
       & \textbf{ResNet-50} &             50.69 & 52.50 & 51.01 & 51.60 & 51.65 & 52.66 & 54.15 & 52.71 & 54.26 & \textbf{55.53} \\
\midrule
\multirow{2}{*}{\textbf{Flowers}} & \textbf{ResNet-18} &             85.96 & 86.05 & 86.02 & 86.03 & 86.40 & 86.25 & \textbf{86.41} & 86.03 & 85.33 & 84.60 \\
       & \textbf{ResNet-50} &             88.75 & 88.30 & 88.57 & 88.27 & \textbf{88.81} & 88.69 & 88.70 & 88.37 & 88.67 & 87.83 \\
\midrule
\multirow{2}{*}{\textbf{Food}} & \textbf{ResNet-18} &             71.77 & 71.83 & 71.73 & 71.64 & 71.60 & 71.64 & \textbf{72.10} & 71.63 & 71.78 & 71.37 \\
       & \textbf{ResNet-50} &             75.83 & 75.19 & 75.52 & 75.51 & 75.50 & 75.37 & \textbf{76.11} & 75.91 & 75.76 & 75.61 \\
\midrule
\multirow{2}{*}{\textbf{Pets}} & \textbf{ResNet-18} &             76.32 & 77.35 & 77.71 & 78.05 & 78.63 & 78.70 & \textbf{78.75} & 77.82 & 75.72 & 72.21 \\
       & \textbf{ResNet-50} &             82.34 & 81.95 & 82.64 & 82.24 & 82.52 & 83.59 & 83.57 & \textbf{83.72} & 81.87 & 79.33 \\
\midrule
\multirow{2}{*}{\textbf{SUN397}} & \textbf{ResNet-18} &             42.81 & 42.65 & 43.40 & 43.35 & 44.01 & 44.20 & 44.51 & \textbf{44.61} & 44.31 & 43.54 \\
       & \textbf{ResNet-50} &             44.64 & 44.95 & 44.73 & 45.09 & 45.44 & 45.93 & 46.74 & 47.24 & \textbf{47.47} & 47.15 \\
\bottomrule
\end{tabular}
\end{small}
\end{table}